\documentclass{article}
\usepackage[main,final]{neurips_2026}
\usepackage{microtype}
\usepackage{graphicx}
\usepackage{subcaption}
\usepackage{booktabs}
\usepackage{hyperref}

\usepackage{amsmath}
\usepackage{amssymb}
\usepackage{mathtools}
\usepackage{amsthm}

\usepackage[capitalize,noabbrev]{cleveref}
\AtBeginDocument{%
  \crefformat{section}{\S#2#1#3}%
  \Crefformat{section}{\S#2#1#3}%
  \crefmultiformat{section}{\S#2#1#3}{ and~\S#2#1#3}{, \S#2#1#3}{, and~\S#2#1#3}%
  \Crefmultiformat{section}{\S#2#1#3}{ and~\S#2#1#3}{, \S#2#1#3}{, and~\S#2#1#3}%
  \crefrangeformat{section}{\S#3#1#4--\S#5#2#6}%
  \Crefrangeformat{section}{\S#3#1#4--\S#5#2#6}%
  \crefformat{subsection}{\S#2#1#3}%
  \Crefformat{subsection}{\S#2#1#3}%
  \crefmultiformat{subsection}{\S#2#1#3}{ and~\S#2#1#3}{, \S#2#1#3}{, and~\S#2#1#3}%
  \Crefmultiformat{subsection}{\S#2#1#3}{ and~\S#2#1#3}{, \S#2#1#3}{, and~\S#2#1#3}%
  \crefrangeformat{subsection}{\S#3#1#4--\S#5#2#6}%
  \Crefrangeformat{subsection}{\S#3#1#4--\S#5#2#6}%
  \crefformat{subsubsection}{\S#2#1#3}%
  \Crefformat{subsubsection}{\S#2#1#3}%
  \crefmultiformat{subsubsection}{\S#2#1#3}{ and~\S#2#1#3}{, \S#2#1#3}{, and~\S#2#1#3}%
  \Crefmultiformat{subsubsection}{\S#2#1#3}{ and~\S#2#1#3}{, \S#2#1#3}{, and~\S#2#1#3}%
  \crefrangeformat{subsubsection}{\S#3#1#4--\S#5#2#6}%
  \Crefrangeformat{subsubsection}{\S#3#1#4--\S#5#2#6}%
  \crefformat{appendix}{\S#2#1#3}%
  \Crefformat{appendix}{\S#2#1#3}%
  \crefmultiformat{appendix}{\S#2#1#3}{ and~\S#2#1#3}{, \S#2#1#3}{, and~\S#2#1#3}%
  \Crefmultiformat{appendix}{\S#2#1#3}{ and~\S#2#1#3}{, \S#2#1#3}{, and~\S#2#1#3}%
  \crefrangeformat{appendix}{\S#3#1#4--\S#5#2#6}%
  \Crefrangeformat{appendix}{\S#3#1#4--\S#5#2#6}%
}

\newtheorem{mytheoremstep}{TheoremProofStep}
\theoremstyle{definition}
\theoremstyle{remark}

\newtheorem*{theorem*}{Theorem}

\usepackage[textsize=tiny]{todonotes}
\graphicspath{ {./images/} }
\usepackage{adjustbox}
\usepackage{multirow}
\usepackage{wrapfig}
\usepackage{scalerel}
\usepackage{float}
\usepackage{fancyhdr}
\usepackage{natbib}
\usepackage{graphicx}
\usepackage{subcaption}
\usepackage{amsthm}
\usepackage{thmtools} 
\usepackage{thm-restate}

\usepackage[T1]{fontenc}    % use 8-bit T1 fonts

\usepackage[utf8]{inputenc}

\usepackage{inconsolata}

\usepackage[scaled=0.85]{helvet}

\usepackage{xltabular}
\usepackage{longtable}
\usepackage{booktabs}       % professional-quality tables
\usepackage{colortbl}
\usepackage{lscape} % for landscape orientation
\usepackage{times}
\usepackage{latexsym}
\usepackage{datetime}       % dates
\usepackage{nicefrac}       % compact symbols for 1/2, etc.
\usepackage{url}            % simple URL typesetting
\usepackage{fnpct}           % nicer footnotes
\usepackage{enumitem}        % nicer itemize
\usepackage{tcolorbox}       % colored background
\usepackage{csquotes}        % quotation marks
\usepackage{xspace}          % adds whitespace after macros
\definecolor{Gray}{gray}{0.9}
\usepackage[dvipsnames]{xcolor}         % colors

\makeatletter
\xspaceaddexceptions{\csq@qclose@i}
\makeatletter

\usepackage{bbm}
\usepackage{amsfonts}
\usepackage{amssymb}
\usepackage{mathtools}
\usepackage{amsmath}
\usepackage{bm}
\usepackage{nicefrac}

\theoremstyle{plain}
\newtheorem{definition}{Definition}

\usepackage{algorithm}
\usepackage{algorithmic}
\makeatother

\usepackage{cleveref}
\crefformat{section}{Section #2#1#3}
\crefname{figure}{Fig.}{Fig.}
\crefname{algorithm}{Alg.}{Alg.}
\crefname{line}{line}{lines}
\crefname{appendix}{App.}{}
\crefname{equation}{eq.}{eqs.}
\crefname{table}{Table}{Tables}
\crefname{proposition}{Proposition}{Propositions}
\crefname{assumption}{Assump.}{Assumps.}
\crefname{definition}{Definition}{Definitions}
\crefname{task}{Task}{Tasks}
\crefname{model}{DNN}{DNNs}
\crefname{submodule}{Submodule}{Submodules}
\crefname{causalgraph}{Alg.}{Alg.}
\crefname{thm}{Theorem}{Theorems}
\crefname{lemma}{Lemma}{Lemmas}

\usepackage[table-column-type=s]{siunitx}

\DeclareSIUnit\thousand{k}
\DeclareSIUnit\million{M}
\DeclareSIUnit\billion{B}
\DeclareSIUnit\trillion{T}
\DeclareSIUnit\x{x}
\DeclareSIUnit\percent{\%}
\DeclareSIUnit\hour{h}
\DeclareSIUnit\min{m}
\DeclareSIUnit\sec{s}

\xspaceaddexceptions{\textsubscript}

\usepackage{tabularray}
\usepackage{cellspace}
\newcommand\cincludegraphics[2][]{\raisebox{-0.2\height}{\includegraphics[#1]{#2}}}
\usepackage{setspace}

\makeatletter
\DeclareRobustCommand*{\escapeus}[1]{%
  \begingroup\@activeus\scantokens{#1 }\endgroup}
\begingroup\lccode`\~=`\_\relax
   \lowercase{\endgroup\def\@activeus{\catcode`\_=\active \let~\_}}
\makeatother

\usepackage[textsize=tiny]{todonotes}

\newcommand{\defn}[1]{\textbf{#1}}

\newcommand{\colourdnn}{purple}
\newcommand{\colourcg}{JungleGreen}
\newcommand{\mydnnmacro}[2]{\newcommand{#1}{{\color{\colourdnn}#2}}}
\newcommand{\mycgmacro}[2]{\newcommand{#1}{{\color{\colourcg}#2}}}

\usepackage{examples-slim}
\usepackage{xfrac}

\renewcommand{\theexample}{\arabic{example}}

\renewcommand{\thesubexample}{\theexample\alph{subexample}}

\makeatletter
\def\@examples{%
  \list{]}{%
    \leftmargin=\exampleindent
    \labelwidth=\exampleindent
    \advance\labelwidth -\labelsep
    \def\@listctr{example}%        % Changed from equation to example
    \@nmbrlisttrue
    \let\makelabel=\@mkexlabel
  }%
}

\def\@mksubexlabel#1{%
  \if\@itemlabel#1\alph{subexample}.%  % Just show letter, not full number
  \else \def\@tempa{#1}\ifx \@tempa\@empty {}%
    \else #1.%
    \fi \fi \hfil
}
\makeatother

\crefname{example}{Example}{Examples}
\Crefname{example}{Example}{Examples}
\crefname{subexample}{Example}{Examples}
\Crefname{subexample}{Example}{Examples}

\crefname{equation}{eq.}{eqs.}
\Crefname{equation}{Equation}{Equations}

\newcommand{\pa}{\textsc{Parasitic Agreement}\xspace}
\newcommand{\co}{\textsc{Carry-One}\xspace}

\newcommand{\prob}{p}
\mycgmacro{\alg}{\mathtt{A}}
\mycgmacro{\algfuncbase}{f}
\mycgmacro{\algfunc}{\algfuncbase_{\alg}}
\newcommand{\algfuncnode}[1]{\algfuncbase_{\alg}^{#1}}

\newcommand{\nodesspace}{\mathcal{N}}

\mycgmacro{\nodevalue}{v}
\mycgmacro{\nodevalueset}{\mathcal{V}}
\mycgmacro{\node}{\eta}
\mycgmacro{\nodes}{\boldsymbol{\eta}}
\newcommand{\nodesinput}{\nodes_{\mathbf{x}}}

\newcommand{\nodesoutput}{\nodes_{\by}}

\newcommand{\nodesint}{\nodes_{\mathtt{int}}}
\newcommand{\nodeint}{\node_{\mathtt{int}}}

\newcommand{\RationalNumbers}{\mathbb{R}}

\mydnnmacro{\dnn}{\mathtt{N}}
\newcommand{\hiddenspace}{\mathcal{H}}
\newcommand{\func}{f}
\mydnnmacro{\dnnfuncbase}{\func}
\mydnnmacro{\dnnfunc}{\dnnfuncbase_{\dnn}}
\newcommand{\dnnfunclayer}[1]{\dnnfuncbase_{\dnn}^{#1}}

\newcommand{\layer}{\ell}
\newcommand{\hiddenstates}{\hiddenstate}
\mydnnmacro{\pdnn}{\prob_{\dnn}}

\mydnnmacro{\neuronset}{\boldsymbol{\psi}}
\newcommand{\neuronsetinput}{\neuronset_{\bx}}
\newcommand{\neuronsetoutput}{\neuronset_{\by}}
\newcommand{\neuronsetint}{\neuronset_{\mathtt{int}}}

\mydnnmacro{\neuron}{\psi}
\mydnnmacro{\Hiddenstates}{\mathbf{H}}

\newcommand{\mathcomment}[1]{\text{\textcolor{gray}{#1}}}

\mydnnmacro{\dnninvfuncbase}{\reflectbox{$f$}}
\newcommand{\matrixidentity}{\mathbf{I}}

\mydnnmacro{\hiddenspacefull}{\boldsymbol{\hiddenspace}}
\mycgmacro{\nodesspacefull}{\boldsymbol{\nodesspace}}

\mycgmacro{\uselesspartition}{\!\bot}

\newcommand{\R}{\RationalNumbers}

\usepackage[utf8]{inputenc} % allow utf-8 input
\usepackage[T1]{fontenc}    % use 8-bit T1 fonts
\usepackage{hyperref}       % hyperlinks
\usepackage{url}            % simple URL typesetting
\usepackage{booktabs}       % professional-quality tables
\usepackage{amsfonts}       % blackboard math symbols
\usepackage{nicefrac}       % compact symbols for 1/2, etc.
\usepackage{microtype}      % microtypography
\usepackage{xcolor}         % colors

\definecolor{darkblue}{rgb}{0, 0, 0.5}
\hypersetup{colorlinks=true, citecolor=darkblue, linkcolor=darkblue, urlcolor=darkblue}

\mydnnmacro{\hiddenstate}{\mathbf{h}}

\newcommand{\Linear}{\calV^{\mathtt{lin}}}
\newcommand{\Rotation}{\calV^{\mathtt{rot}}}

\newcommand{\RotationLinear}{\calV^{\mathtt{r+l}}}

\newcommand{\rotationlinearname}{rotation+linear\xspace}

\newcommand{\calX}{\mathcal{X}}
\newcommand{\bx}{\mathbf{x}}
\newcommand{\bxbase}{\bx^{\mathtt{b}}}
\newcommand{\bxsrc}{\bx^{\mathtt{s}}}
\newcommand{\by}{\mathbf{y}}
\newcommand{\bybasesrc}{\by^{\mathtt{s}\rightarrow\mathtt{b}}}
\newcommand{\bybasesrcA}{\bybasesrc_\alg}
\newcommand{\bybasesrcN}{\bybasesrc_\dnn}

\newcommand{\calY}{\mathcal{Y}}
\newcommand{\nodesrc}{\nodevalue^{\mathtt{s}}}
\newcommand{\nodesrcbase}{\nodevalue^{\mathtt{s}\rightarrow\mathtt{b}}}
\newcommand{\hiddenbase}{\hiddenstate^{\mathtt{b}}}
\newcommand{\hiddensrc}{\hiddenstate^{\mathtt{s}}}
\newcommand{\hiddensrcbase}{\hiddenstate^{\mathtt{s}\rightarrow\mathtt{b}}}
\newcommand{\latent}{\hiddenstate_{\causalmap}}
\newcommand{\latentonesrc}{\hiddenstate^{\mathtt{s}}_{\causalmap,1}}
\newcommand{\latenttwobase}{\hiddenstate^{\mathtt{b}}_{\causalmap,2}}
\newcommand{\latentdbase}{\hiddenstate^{\mathtt{b}}_{\causalmap,d}}
\newcommand{\latenti}{\hiddenstate_{\causalmap,i}}
\newcommand{\latentsrcbase}{\latent^{\mathtt{s}\rightarrow\mathtt{b}}}

\newcommand{\calV}{\mathcal{V}}

\newcommand{\makesf}[1]{\textsf{{\escapeus{#1}}}}

\title{Parasite Features: Causal Abstraction Through Spurious Pathways}

\newcommand{\ethzid}{1}
\newcommand{\utaustinid}{2}
\newcommand{\epflid}{3}

\author{%
  Denis Sutter,\thanks{Equal contribution.}$^{\,\,\,,\ethzid}$ Qing Yao,$^{*,\utaustinid}$ Sasha Boguraev,$^{*,\utaustinid}$ Julian Minder,$^\epflid$ Antonio Di Maio,$^\ethzid$ \\
  \textbf{Thomas Hofmann},$^\ethzid$ \textbf{Kyle Mahowald},$^\utaustinid$ \textbf{Tiago Pimentel}$^\ethzid$ \\
  $^\ethzid$ ETH Zurich,\quad
  $^\utaustinid$ The University of Texas at Austin,\quad
  $^\epflid$ EPFL
  \\
  \texttt{densutter@ethz.ch},\quad
  \texttt{sasha.boguraev@utexas.edu}, \\
  \texttt{qyao@utexas.edu},\quad
  \texttt{tiago.pimentel@inf.ethz.ch} 
  \\[5pt]
  \cincludegraphics[width=1.2em, keepaspectratio]{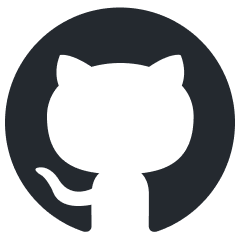}
     \href{https://github.com/dounick/parasite-features}{{\makesf{dounick/parasite-features}}}
}

\begin{document}

\maketitle
\vspace{-15pt}

\begin{abstract}
Causal Abstraction defines a notion of \emph{equivalence} between a neural network and an algorithm, thus serving as a powerful framework for mechanistic interpretability. In this paper, however, we show the existence of a class of \emph{spurious features} which straightforwardly pass such equivalence tests. Suppose there are two directions in a model's representation space where: (i) one discriminates the input based on some feature, and (ii) the other steers the model towards a certain output. Interpretability methods based on Causal Abstraction (e.g., Distributed Alignment Search) can \emph{spuriously} simulate these directions' combined behavior by leveraging pathways passing through them, giving the impression that the model implements this feature.
In such cases, we say the method discovered a \defn{parasite feature}. 
Experimentally, we show that Distributed Alignment Search finds parasite features using linear alignment maps in two tasks: syntactic agreement and two-digit addition. Our paper highlights an important consideration for Causal Abstraction and contributes to the ongoing debate about best practices for mechanistic interpretability.
\end{abstract}

\vspace{-5pt}
\section{Introduction}
\vspace{-2pt}

\newcommand{\causalmap}{\boldsymbol{\phi}}

The field of mechanistic interpretability aims to uncover the mechanisms underlying a model's inner workings \citep{olah2020zoom}, encompassing many methods and frameworks \citep[e.g.][]{elhage2021mathematical,mueller2024questrightmediatorhistory,ferrando2024primerinnerworkingstransformerbased,sharkey2025openproblemsmechanisticinterpretability}. 
Among these, \defn{Causal Abstraction} is particularly promising, formally defining notions of equivalence between causal graphs \citep{Beckers2018AbstractingCM}.
Under this framework, a deep neural network (DNN) is equivalent to a (conceptually simpler) algorithm if they behave similarly under interventions \citep{geiger2025causal}; whenever this is the case, we informally say that the DNN \emph{implements} the algorithm.
Causal Abstraction thus allows us to interpret a DNN in terms of the algorithms it implements.\looseness=-1

In recent years, however, a number of works have raised potential pitfalls in using mechanistic interpretability approaches, including Causal Abstraction, to draw firm conclusions about a model \citep[e.g.,][\textit{inter alia}]{Mueller2024MissedCA,sutter2025nonlinearrepresentationdilemmacausal,meloux2025dead}.
Most relevant to us are two concerns.
First, mechanistic interpretability approaches may be \emph{non-identifiable} and a DNN might be found to implement many algorithms \citep{meloux2025everything}.
Second, Causal Abstraction methods may find \emph{interpretability illusions}, i.e., interventions that leverage a DNN's null space to imitate genuine causal behavior \citep{Makelov2023IsTT}. 
As noted by \citet{wu2024replymakelovetal} in their response to \citeauthor{Makelov2023IsTT}, though, interpretability illusions may not pose an actual obstacle for Causal Abstraction, as it is not inherently problematic for an intervention to leverage a model's null space.\looseness=-1

\begin{figure}[t]
    \centering
    \includegraphics[width=.95\linewidth]{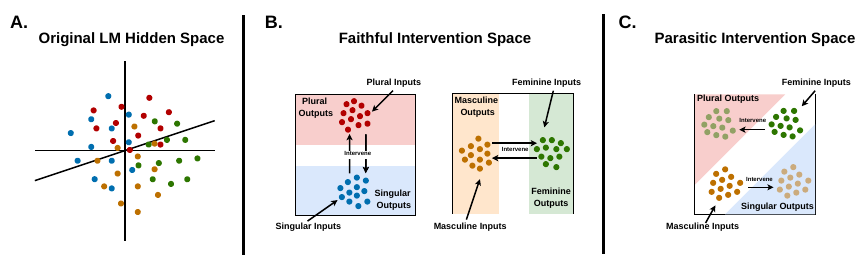}
    \caption{A schematic of parasite features for a hypothesized case of syntactic agreement (described in \cref{sec:motivating}). \textbf{A:} Prompts in the original high-dimensional DNN hidden space are not basis-aligned by feature. \textbf{B:} By projecting onto subspaces, inputs can be separated by features. Interventions in this space move inputs to the region with corresponding output behavior. \textbf{C:} However, a linear alignment map may connect causally unrelated directions. Interventions in this space create the illusion of causality, despite there being no such relationship in the model. This depicts a \emph{parasite feature.}\looseness=-1}
    \label{fig:fig1}
    \vspace{-5pt}
\end{figure}

In this paper, we identify a related set of interpretability illusions  which masquerade as causal features by manipulating spurious pathways in a DNN. 
Under mild assumptions, we show that alignment search methods can discover \defn{parasite features}, a mechanistic interpretation of model behavior which \emph{spuriously} mimics a target feature by leveraging two directions in the model where: (i) one discriminates the inputs in terms of this feature, and (ii) another steers the model's output towards an intended output.
With the right alignment map and hidden state geometry, these directions can be linked together, leading to seemingly causal behavior under intervention, despite the leveraged directions being (under normal operation) causally unrelated in the model (see \cref{fig:fig1} for a visual example).\looseness=-1

We present a theoretical account of parasite features (\cref{sec:def,sec:theoretical_setup}), followed by two empirical case-studies showing such parasite features are not merely theoretical concerns in toy models, but a practical concern (\cref{sec:set-up,sec:results}). 
Concretely, we find parasite features in two tasks, syntactic agreement and two-digit addition, by training interventions with a constrained class of linear alignment maps (which are minimally more expressive than rotation maps).
We end with a discussion on the prevalence of parasite features and what these results mean for interpretability research more broadly (\cref{sec:discussion}).\looseness=-1

Our work demonstrates that while matching an algorithm's behavior under intervention is necessary for a feature to be faithful, it is not sufficient, even when alignments are restricted to linear or rotation maps.
Claims that rest on high intervention accuracy alone should thus be treated with care, with our results further highlighting the importance of sound scientific practice in interpreting the results of mechanistic interpretability experiments.
In doing so, we join a growing chorus emphasizing careful interpretation of mechanistic results, with a focus on robustness and generalizability \citep{geiger2025causalabstractionunderpinscomputational,sutter2025nonlinearrepresentationdilemmacausal, meloux2025dead, potts2025interp}.\looseness=-1

\vspace{-3pt}
\section{Causal Abstraction}\label{sec:ca}
\vspace{-2pt}

In this section, we introduce relevant aspects of the Causal Abstraction framework \citep{rubenstein2017causal, Beckers2018AbstractingCM} used throughout our work.
Causal Abstraction defines notions of equivalence between two structural causal models.
In our case, these causal models are a deep neural network (DNN) $\dnn$ and an algorithm $\alg$ which we might hypothesize the DNN implements.

First, we represent the \defn{algorithm} $\alg$ as an acyclic directed causal graph, which contains: 
(i) input nodes $\nodesinput$ representing input elements $\bx \in \calX$,
(ii) output nodes $\nodesoutput$ representing elements $\by \in \calY$, 
and (iii) inner nodes $\nodesint$ representing intermediate variables whose value is computed as a function of its parents' values.\footnote{For more general formalisms covering, e.g., cyclic causal models, see \citet{Beckers2018AbstractingCM} or \citet{geiger2025causal}.\looseness=-1}
For simplicity of exposition, we will focus on analyses with a single feature, and we thus consider this graph to have a single inner node $\nodeint$.
Our results, however, can be easily extended to more complex graphs. 
Algorithm $\alg$'s computation can be summarized as: $\nodevalue = \algfuncnode{\nodeint}(\bx)$ and $\by = \algfuncnode{\nodesoutput}(\bx, \nodevalue)$.
Now, consider this algorithm's behavior $\alg$ under two different inputs, which we will call the \textbf{base} ($\bxbase$) and the \textbf{source} ($\bxsrc$).
We define a \defn{source-to-base intervention} on inner node $\nodeint$ as replacing this node's value in the base forward pass by its value under the source input,
and continuing its forward pass as usual.
Effectively, the intervention replaces just the feature value represented by $\nodeint$ in the base with a counterfactual value in its computation, leaving everything else untouched:\looseness=-1
\begin{align}
    \nodesrc = \algfuncnode{\nodeint}(\bxsrc),
    \qquad\quad
    \nodesrcbase = \nodesrc,
    \qquad\quad
    \bybasesrcA = \algfuncnode{\nodesoutput}(\bxbase, \nodesrcbase)
\end{align}

Many \defn{DNNs} $\dnn$ can be similarly represented as acyclic directed causal graphs.
In the case of transformer LMs, we can focus on a single layer $\layer$ and position, and write: $\hiddenstate = \dnnfunclayer{\layer}(\bx)$ and $\by = \dnnfunclayer{\by}(\hiddenstate)$, where $\hiddenstate \in \R^d$ is the hidden state.
Representing each neuron in this layer as a distinct node, $\dnn$ is thus a causal graph with nodes $\neuronsetinput \cup \neuronsetoutput \cup \neuronsetint$, where $\neuronsetinput$, $\neuronsetoutput$ are defined as for the algorithm, and $\neuronsetint$ contains $d$ nodes corresponding to neurons in $\hiddenstate$.
We can now define source-to-base interventions on a DNN as before: given a chosen neuron $i$, compute hidden states for both the base and source, 
replace the base value of a neuron $i$ with the source's,
and continue the forward pass. Formally:
% ($\by = \dnnfunclayer{\by}(\hiddensrcbase)$).
\begin{equation}\label{eq:intervention_dnn}
    \hiddenbase = \dnnfunclayer{\layer}(\bxbase),
    \quad
    \hiddensrc = \dnnfunclayer{\layer}(\bxsrc),
    \quad 
    \hiddensrcbase = \big(\hiddenbase_1,\hiddenbase_2, \dots, \hiddensrc_i, \dots, \hiddenbase_d \big),
    \quad
    \bybasesrcN = \dnnfunclayer{\by}(\hiddensrcbase)
\end{equation}

Several definitions of equivalence between causal graphs fall under Causal Abstraction.
\defn{Constructive abstraction} \citep{Beckers2018AbstractingCM}, defines an equivalence between $\dnn$ and $\alg$ iff there exists a set of neurons such that $\bybasesrcN = \bybasesrcA$ holds for any $\bxsrc$, $\bxbase$ pair.\footnote{
More precisely, constructive abstraction \citep{Beckers2018AbstractingCM} requires two stronger conditions to hold: 
(i) $\bybasesrcN = \bybasesrcA$ must hold for any ``\emph{compatible}'' intervention-pair ($\hiddensrcbase$, $\nodesrcbase$) not just pairs generated as above;
(ii) the intervened value $\hiddensrc_i$ must encode sufficient information to predict $\nodesrc$.
Given transformer's injectivity property \citep{sutter2025nonlinearrepresentationdilemmacausal,nikolaou2026languagemodelsinjectiveinvertible}, we assume (ii) holds here.
Regarding (i), we will work here with the input-restricted variants of causal abstraction \citep[presented by][]{sutter2025nonlinearrepresentationdilemmacausal}, thus softening requirement (i) to intervention-pairs ($\hiddensrcbase$, $\nodesrcbase$) defined as above.
See \citet{sutter2025nonlinearrepresentationdilemmacausal} for a longer discussion on these definitions, and a more precise definition of ``compatible'' interventions.\looseness=-1}
This definition may be too restrictive, as it assumes the high-level conceptual information represented by $\nodeint$ is encoded in neuron-aligned bases of the representation space---something typically deemed too strong in DNNs (see \citealp{sutter2025nonlinearrepresentationdilemmacausal} for a longer discussion).
In turn, other notions of Causal Abstraction \citep{geiger2023findingAB,geiger2025causal,sutter2025nonlinearrepresentationdilemmacausal} define an equivalence between $\dnn$ and $\alg$ allowing for information to be latently encoded in $\hiddenstate$.
Formally, let $\causalmap: \R^d \to \R^d$ be an \defn{alignment map}: a learned bijective function which exposes a set of latent variables the representations $\hiddenstate$ encode, with $\latent = \causalmap(\hiddenstate)$.
We can use such an alignment map to perform source-to-base interventions on $\latent$ as in \cref{eq:intervention_dnn}, producing vector $\latentsrcbase$.
The notion of \defn{$\boldsymbol{\calV}$-abstraction} then defines an equivalence between $\dnn$ and $\alg$ if there exists a $\causalmap \in \calV$ (for a variational family of bijective functions $\calV$) such that $\bybasesrcN = \bybasesrcA$ holds for any $\bxsrc$, $\bxbase$ pair.

The chosen class of alignment maps $\calV$ reflects assumptions about how DNNs encode conceptual information.
Under the linear representation hypothesis \citep{alain2018understanding}, this class would be restricted to rotations \citep{geiger2023findingAB}, or the more expressive class of invertible linear maps \citep{grant2025emergentsymbollikenumbervariables,grant2025repdivergence}.
However, there is growing evidence that neural networks encode relevant feature information nonlinearly 
\citep{white-etal-2021-non,pimentel-etal-2022-attentional,csordas-etal-2024-recurrent,levy-geva-2025-language,kantamneni2025languagemodelsusetrigonometry,mib-2025}, which warrants the use of nonlinear maps.
At the same time, increasing $\calV$'s expressivity has risks: if $\calV$ encompasses arbitrary nonlinear bijections, \emph{any} neural network can be mapped to \emph{any} algorithm \citep{sutter2025nonlinearrepresentationdilemmacausal}. In this work, we show even rotation or linear alignment maps are not safe from fitting spurious features.

\vspace{-3pt}
\section{Motivating Example: Parasitic Agreement}\label{sec:motivating}
\vspace{-2pt}

We start with a thought experiment to motivate parasite features focusing on two forms of syntactic agreement: gender-agreement and number-agreement. An example of gender-agreement can be seen in \Cref{ex:gender-agreement}. Here, the grammatical gender of the sentence's subject (as in \textit{\textcolor{BurntOrange}{man}} vs.\ \textit{\textcolor{ForestGreen}{woman}}) is responsible for licensing the correct anaphoric pronoun (\textit{\textcolor{BurntOrange}{he}} vs.\ \textit{\textcolor{ForestGreen}{she}}).

\vspace{-5pt}
\begin{examples}
  \item
    {
    \begin{examples}
      \setlength{\exampleindent}{1em}
      \setlength{\labelsep}{0.2em}
      \item The \textbf{\textcolor{BurntOrange}{man}} ran down the street because $\Rightarrow$ \textbf{\textcolor{BurntOrange}{he}}
      \item The \textbf{\textcolor{ForestGreen}{woman}} ran down the street because $\Rightarrow$ \textbf{\textcolor{ForestGreen}{she}}
    \end{examples}
    }\label{ex:gender-agreement}
\end{examples}
\vspace{-5pt}

Similarly, in the case of number-agreement (\cref{ex:number-agreement}), the plurality of the sentence's subject (as in \textit{\textcolor{blue}{key}} vs.\ \textit{\textcolor{red}{keys}}) is responsible for licensing the correct verb form (\textit{\textcolor{blue}{is}} vs.\ \textit{\textcolor{red}{are}}).

\vspace{-5pt}
\begin{examples}
  \item
    {
    \begin{examples}
      \setlength{\exampleindent}{1em}
      \setlength{\labelsep}{0.2em}
      \item The \textbf{\textcolor{blue}{key}} next to the counter $\Rightarrow$ \textbf{\textcolor{blue}{is}}
      \item The \textbf{\textcolor{red}{keys}} next to the counter $\Rightarrow$ \textbf{\textcolor{red}{are}}
    \end{examples}
    }\label{ex:number-agreement}
\end{examples}
\vspace{-5pt}

A competent LM should recognize these dependencies and predict the correct output given the upstream item. 
One should thus be able to learn an alignment map such that manipulating the upstream feature in an LM leads to a corresponding change in the LM's output behavior. 

However, let us now consider a manufactured example in which the gender of a sentence's subject is responsible for licensing the plurality of a downstream verb (\cref{ex:parasite-agreement}).

\vspace{-5pt}
\begin{examples}
  \item
    {
    \begin{examples}
      \setlength{\exampleindent}{1em}
      \setlength{\labelsep}{0.2em}
      \item The \textbf{\textcolor{BurntOrange}{man}} next to the counter $\Rightarrow$ \textbf{\textcolor{blue}{is}}
      \item The \textbf{\textcolor{ForestGreen}{woman}} next to the counter $\Rightarrow$ \textbf{\textcolor{red}{are}}
    \end{examples}
    }\label{ex:parasite-agreement}
\end{examples}
\vspace{-5pt}

We would expect our linguistically-competent LM to not behave accordingly. As such, there should not be an alignment map under which manipulating subject gender controls downstream verb plurality. But this is not guaranteed, and below we construct a simple counterexample.

Consider a normal English LM, $\dnn$, that performs number-agreement. Assume that in some layer $\layer$, there exists a direction that influences downstream verb number.
We can construct another LM, $\dnn'$, which is an exact copy of $\dnn$, except for an additional neuron in layer $\layer$. 
This neuron is computationally inert, having no outgoing weights, and simply encodes gender (e.g., with activation $-1$ for masculine, $+1$ for feminine).
By construction, $\dnn'$ is behaviorally indistinguishable from $\dnn$ on all inputs.

However, despite the additional neuron playing no causal role in the model, it can be exploited by alignment maps. 
Consider a class of maps $\calV$ and an alignment map $\causalmap \in \calV$ that projects the model's activations onto a direction at a $45^{\circ}$ angle with: (i) the direction of the extra inert neuron and (ii) the direction causally associated with \textbf{number}---virtually combining them into a latent feature.
By linking these directions, $\causalmap$ can induce changes in downstream verb plurality, despite the signal originating from a feature (\textbf{gender}) which is never used by our model (as is exemplified in \cref{fig:fig1}C).

This is an archetypal parasite feature --- a counterfactual gender-to-number feature is assumed to be implemented in $\dnn$ by an alignment map which combines unrelated directions in a manner which is irrelevant to the model's computation.
In real LMs, there are no artificial neurons that encode input features, but such directions often natively exist (e.g., a direction which spuriously separates inputs by gender in a model that performs number-agreement; \citealp{ravichander-etal-2021-probing}).
While these directions are not as explicitly causally inert as the added neuron in $\dnn'$, there is likewise no guarantee that interventions leveraging them are faithful to the model's computation.\footnote{We note that such an example is very similar to the one provided by \citet{Makelov2023IsTT}. Our example differs, however, in a couple regards. First, our steering direction is the LM's \emph{natively active, on-manifold} direction, whereas Makelov's ``dormant'' steer direction is zero on-data and only fires off-distribution; second, we link \emph{two distinct concepts} rather than a disconnected correlate of the \emph{same} feature. For a larger discussion on the similarities and differences between the phenomena see \cref{sec:discussion}.\looseness=-1}

\newcommand{\directionvector}{\boldsymbol{\neuron}}
\newcommand{\featuredeterminingdirection}{\directionvector^{\mathtt{DD}}}
\newcommand{\outputinfluencedirection}{\directionvector^{\mathtt{SD}}}

\vspace{-3pt}
\section{Defining Features, Faithful Features, and Parasite Features}\label{sec:def}
\vspace{-2pt}

Before formalizing \emph{parasite features}, we will first formalize the concept of a \emph{feature}. We take the perspective that a \defn{feature} is a node in a causal graph.
In this sense, our \defn{ground truth features} are elements in $\nodeint$.
This has at least one important implication:
a feature has not only an \defn{input-side component}, which defines the set of values it takes given the input (i.e., $\nodevalue = \algfuncnode{\nodeint}(\bx)$),
but also an \defn{output-side component}, which defines its impact on an algorithm's outputs (i.e., $\by = \algfuncnode{\nodesoutput}(\bx, \nodevalue)$).
We also have \defn{candidate features} in a DNN: each of its explicit $\hiddenstate_i$ or latent $\latenti$ nodes is a potential feature used for computations.
Our goal is then to find features in a DNN which are faithful to a target feature.\looseness=-1

Given a target feature ($\nodeint$), we can use different techniques to find it in a model (i.e., to find an equivalent $\latenti$), including diagnostic probing \citep{alain2018understanding}, sparse-autoencoders \citep{bricken2023monosemanticity, huben2024sparse}, and distributed alignment search \citep{geiger2023findingAB}.
However, how do we know the feature we find in a model is faithful?
\citet{jacovi-goldberg-2020-towards} intuitively define a \defn{faithful feature} to be one that \emph{accurately represents the reasoning process behind the model’s prediction} \citep[although see][\textit{inter alia}, for more definitions]{10.1145/2939672.2939778, wiegreffe-pinter-2019-attention, bastings-filippova-2020-elephant, hanna2024have}.\footnote{Faithfulness is thus a property of a candidate feature as compared to a ground truth, being a property of a triple: $\langle \alg, \dnn, \causalmap\rangle$.}
Under this definition, a clear condition for a feature's faithfulness is being \defn{causally meaningful}: i.e., changes to the candidate feature in $\dnn$ must match the ground truth feature's behavior under $\alg$.
Importantly, causal abstraction gives us---by construction---features which are causally meaningful in this sense, as interventions on the model $\dnn$ must match the evaluated algorithm $\alg$.

Is causal meaningfulness a sufficient condition for a feature to be faithful, however? We argue that it is not.
\citet{sutter2025nonlinearrepresentationdilemmacausal} showed that, given a strong enough class of alignment maps $\calV$, we can find \emph{any} feature in \emph{any} model using causal abstraction.
In this paper, we document another class of features which we argue are not faithful, but still pass the causally meaningful test: parasite features.

\newcommand{\inputfunc}{f}
\newcommand{\outputfunc}{g}
\newcommand{\calH}{\mathcal{H}}

\newcommand{\steeringgap}{\varepsilon_{\mathtt{gap}}}
\newcommand{\steeringmax}{\varepsilon_{\mathtt{max}}}
\newcommand{\steeringthreshold}{\delta_{\mathtt{steer}}}
\newcommand{\discriminationthreshold}{\delta_{\mathtt{disc}}}
\newcommand{\classspread}{\varepsilon_{\mathtt{spread}}}

Parasite features are, informally, a spurious connection between two directions of a DNN's hidden space: a discriminative direction $\featuredeterminingdirection$, which separates hidden states into two distinct clusters; and a steering direction $\outputinfluencedirection$, which predictably affects the model's output behavior when perturbed. 
Formally, let $\inputfunc: \calX \to \{0,1\}$ be an input-feature function and $\outputfunc: \calX \times \{0,1\} \to \calY$ be an output-behavior function.
Now, let an input $\bx$ be \defn{measurably true in input-feature} $\inputfunc$ if $\inputfunc(\bx)=1$, and measurably false otherwise; 
we can use this definition to cluster $\dnn$'s hidden states (i.e., $\calH = \{\dnnfunclayer{\layer}(\bx) \mid \bx \in \calX\}$) into sets $\calH_0$ and $\calH_1$ produced by measurably false and true inputs, respectively.
Further, let an input--output pair $\bx,\by$ be \defn{measurably true in output-behavior} $\outputfunc$ if $\by = \outputfunc(\bx, 1)$, measurably false if $\by = \outputfunc(\bx, 0)$, and undefined otherwise.\footnote{Note that, given a binary feature $\nodeint$, our definitions of $\inputfunc$ and $\outputfunc$ precisely match an algorithm's structure $\nodevalue = \algfuncnode{\nodeint}(\bx)$ and $\by = \algfuncnode{\nodesoutput}(\bx, \nodevalue)$. We make this connection more precise later.
Further note that the definitions of `measurably true' and `measurably false' in $\outputfunc$ can be trivially extended to operate on $\dnn$'s logits directly (e.g., the probability of a True class should exceed a False class), instead of on the produced output $\by$. This fact allows us to treat the logit-based metrics utilized in \cref{sec:results} as ``measurably true/false'', allowing experimental validation of our theoretical proof.
}
We can now define directions $\featuredeterminingdirection$ and 
$\outputinfluencedirection$ formally.\looseness=-1

\begin{definition}
    \label{def:discriminative_direction}
    Let $\dnn$ be a DNN and $\inputfunc$ an input-feature function. 
    A unit vector $\featuredeterminingdirection \in \R^d$ is a \defn{discriminative direction} for $(\dnn, \inputfunc)$ if there exists a threshold $\discriminationthreshold > 0$ which separates the prompts into clusters such that: $\langle \featuredeterminingdirection, \hiddenstate \rangle > \discriminationthreshold$ for all $\hiddenstate \in \mathcal{H}_1$ and $\langle \featuredeterminingdirection, \hiddenstate \rangle < -\discriminationthreshold$ for all $\hiddenstate \in \mathcal{H}_0$.
\end{definition}

\begin{definition}
\label{def:steering_direction}
Let $\dnn$ be a DNN and $\outputfunc$ an output-behavior function.
A unit vector $\outputinfluencedirection \in \R^d$ is a \defn{steering direction} for $(\dnn, \outputfunc)$
if there exists a threshold $\steeringthreshold > 0$ such that whenever $\langle \hiddenstate, \outputinfluencedirection \rangle > \steeringthreshold$, the output of $\hiddenstate$ under $\dnn$ is \textbf{measurably true} in behavior $\outputfunc$, and \textbf{measurably false} whenever $\langle \hiddenstate, \outputinfluencedirection \rangle < -\steeringthreshold$.\looseness=-1
\end{definition}

Our definition of steering directions is intentionally broad. Some steering directions are intuitive: a model can implement a direction that \emph{discriminates} inputs by grammatical number and that also \emph{steers} its outputs to change a verb's number.
However, some steering directions can be more abstract: e.g., a model can have a direction that systematically adds a constant to arithmetic results, or that steers an output verb's number without simultaneously discriminating inputs.
Strange though we know these directions may seem, we find that they exist in pretrained models (in \Cref{sec:results}; with further evidence of their prevalence in \Cref{app:other_parasites}).
These directions allow parasite features to emerge.

\begin{definition}\label{def:parasite_features}
     Consider an $\dnn,\inputfunc, \outputfunc$ with two \emph{causally independent} directions: a discriminative direction, $\featuredeterminingdirection$, and steering direction, $\outputinfluencedirection$. 
     Finally, let $\alg$ have $\algfuncnode{\nodeint} = \inputfunc$ and $\algfuncnode{\nodesoutput} = \outputfunc$.
     We say an alignment map discovers a \textbf{parasite feature} if it leverages a combination of $\featuredeterminingdirection$ and $\outputinfluencedirection$ to show $\dnn$ implements $\alg$, using the discriminative direction $\featuredeterminingdirection$ to move representations in the steering direction $\outputinfluencedirection$.\footnote{Parasitism is a property of a candidate feature as compared to a ground truth, and thus a property of $\langle\alg, \dnn, \causalmap\rangle$.}
\end{definition}

Importantly, we rely informally on the term `causally independent' above to mean the input-side feature $\inputfunc$ and output-side feature $\outputfunc$ `operate independently' within the model. 
Defining what it means for these directions to `operate independently' in a deep neural network, however, is non-trivial. 
Instead, we offer two conditions which we believe should (in general) be sufficient for such independence to hold: \textbf{(i) data-level independence}, in which directions operate on conceptually distinct and causally unrelated features of the training data (e.g., subject gender and verb number), so a competent model should process them independently, and \textbf{(ii) subspace-level independence}, in which the two directions do not interact in the model's computation, meaning that moving along the discriminative direction does not affect the decision boundary set by the steering direction.
We argue only that these conditions are sufficient --- but not strictly necessary --- for two directions to operate independently within a competent model.
Two directions ($\featuredeterminingdirection$ and $\outputinfluencedirection$) which operate on conceptually related features (data-level dependent) may still be processed independently by the model. 
Similarly, two directions may influence one another (subspace-level dependence), but the model may have mechanisms in place to offset those interactions under its normal operation.\looseness=-1

\vspace{-3pt}
\section{Sufficient Conditions for Discovering Parasite Features}
\label{sec:theoretical_setup}
\vspace{-2pt}

\newcommand{\LinearMatrix}{\causalmap}

We now state formal conditions under which there exists an invertible linear map that can leverage a discriminative direction for $\inputfunc$
to influence the model's behavior towards $\outputfunc$ via a steering direction,
regardless of any causal relationship between them. 
The implications of the theorem are particularly worrisome when $f$ and $g$ are causally disconnected in a target domain, making a feature $\nodeint$ combining their behavior implausible.
Now, let $\calH=\{\dnnfunclayer{\layer}(\bx) \mid \bx \in \calX\}$ be a set of $d$-dimensional hidden states at layer $\layer$, with $d\geq 2$, partitioned by $\inputfunc$ into $\calH_0,\calH_1$. We denote elements of $\calH$ by $\hiddenstates$.

We define our theorem's conditions as follows.

\begin{restatable}{assumption}{outputsteerability}
\label{assumption:output}
\textbf{(Steerability.)} 
A steering direction $\outputinfluencedirection$ exists for $\outputfunc$ with threshold $\steeringthreshold>0$.\looseness=-1
\end{restatable}

\begin{restatable}{assumption}{inputdiscrimination}
\label{assumption:input}
    \textbf{(Discrimination.)}
    A discriminative direction $\featuredeterminingdirection$ exists for $\inputfunc$ with threshold $\discriminationthreshold\!>\!0$.\looseness=-1
\end{restatable}

\begin{restatable}{assumption}{sufficientseparation}
\label{assumption:separation}
    \textbf{(Sufficient separation.)} 
    Let 
    (i) $\steeringgap>0$ be the minimum gap between hidden states and the $\pm \steeringthreshold$ steering threshold;
    (ii) $\steeringmax>0$ be the maximum activation of the steering direction by hidden states; and
    (iii) $\classspread > 0$ be the maximum within-class spread along $\featuredeterminingdirection$:
    \begin{equation}
        \steeringgap \!=\!  \min_{\hiddenstate\in\calH}\bigl|\,|\langle\hiddenstate,\outputinfluencedirection\rangle|-\steeringthreshold\,\bigr|,
        \,\,\,\,
        \steeringmax \!=\! \max_{\hiddenstate\in\calH} |\langle\hiddenstate,\outputinfluencedirection\rangle|,
        \,\,\,\,
        \classspread \!=\! \max_{\hiddenstate_s,\hiddenstate_b\in\mathcal{H}_i} |\langle\featuredeterminingdirection,\hiddenstate_s-\hiddenstate_b\rangle|
    \end{equation}
    We require $|\langle\featuredeterminingdirection,\outputinfluencedirection\rangle| < 1$ and a minimum gap-to-spread ratio satisfying $\frac{\steeringgap}{\classspread} > \dfrac{\steeringmax + \steeringthreshold}{2\,\discriminationthreshold}$.
\end{restatable}

In words, assumptions \ref{assumption:output} and \ref{assumption:input} state the functions are associated with discriminative and steering directions in the model, while
assumption \ref{assumption:separation} defines a minimum ratio between the gap between representations and the steering direction; and the spread of representations per cluster $\calH_i$ along the discriminative direction.
As representations naturally cluster according to input features, we expect the mere existence of steering directions to make these conditions reasonable and to hold for a number of $\inputfunc,\outputfunc$ function pairs at some position and layer within a model.

We can now present our theorem.

\begin{restatable}{thm}{generalthm}\label{theorem:general}
Let $\dnn$ be a DNN, and $\alg$ be an algorithm with $\algfuncnode{\nodeint} = \inputfunc$ and $\algfuncnode{\nodesoutput} = \outputfunc$.
Further, let $\Linear$ be the class of invertible linear maps.
Under assumptions 1, 2, and 3, $\alg$ is a $\Linear$-abstraction of $\dnn$.
\end{restatable}
\vspace{-5pt}
\begin{proof}[Proof sketch]
See \cref{app:proof} for a formal proof.
We prove this by construction, building an alignment map $\causalmap \in \Linear$ with the following intuition: given two points in the separate sets $\calH_0,\calH_1$, provided the inputs are cleanly separated and clustered along the discriminative direction, an invertible linear matrix can always rotate the difference between the two points to align with the steering direction, and amplify it appropriately to ``bridge the gap'' in the steering direction. 
Assumption \ref{assumption:separation} further ensures that perturbations for within-class interventions remain in their output region.
\end{proof}

Intuitively, \cref{theorem:general} shows that, if assumptions 1--3 are satisfied, there exists an alignment map $\causalmap \in \Linear$ which aligns $\alg$ to $\dnn$, such that $\dnn$'s output aligns with the behavior of a combined $(\inputfunc,\outputfunc)$.
Notably, this theorem holds even if directions $(\featuredeterminingdirection,\outputinfluencedirection)$ are orthogonal (suggesting subspace-level independence) or operate on conceptually unrelated features (suggesting data-level independence).
We prove a similar theorem, with stricter assumptions, for the class of rotation alignment maps in \Cref{app:rotation_proof}. 
This highlights that parasite features are a general problem for Causal Abstraction, and that even less expressive alignment maps are not necessarily immune to discovering spurious features.

\vspace{-3pt}
\section{Experimental Setup}\label{sec:set-up}
\vspace{-2pt}

We find parasite features in two tasks: first, a nonsensical syntactic task (\pa) involving unrelated syntactic features; second, a task localizing a carry-one feature in two-digit addition (\co).
These tasks differ in that we have strong reason to believe the model should \emph{not} be competent in the syntax task, as it is constructed to be parasitic. In contrast, the carry-one feature in two-digit addition appeared in the mechanistic interpretability benchmark \citep[MIB;][]{mib-2025} as a plausible faithful feature. We discuss additional parasite features and tasks in \Cref{app:other_parasites}.

\vspace{-3pt}
\subsection{Tasks and Causal Algorithms}
\vspace{-2pt}

\paragraph{\pa.} 

We create pairs of single-clause sentences labeled for subject gender. Sentences with masculine subjects are then given a singular verb, whereas feminine subjects have plural verbs (see \cref{ex:parasite-agreement}).
A candidate causal-algorithm, $\alg_\mathtt{agr}$, for solving this task can be informally stated as: if the subject of the sentence is masculine, output a singular verb, otherwise if the subject of the sentence is feminine, output a plural verb.

We use $\texttt{pythia-1b}$ \citep{biderman2023pythia} because it is competent at number and gender agreement, and rotation alignment maps can align these features with strong causal effects \citep{arora-etal-2024-causalgym}.
In relation to \Cref{theorem:general}, the output behavior is whether singular continuations are favored over plurals, with the steering direction controlling logits of singular and plural verbs. All prompts begin in the singular region (subjects are always singular by design), so the alignment map must transport feminine-subject hidden states to the plural region under intervention.

\paragraph{\co.}

Models receive string inputs of a two-digit addition problem and provide the output string of the correct sum (see \cref{ex:carryone-paired}).
The mechanistic interpretability benchmark \citep{mib-2025} hypothesizes a model may use the following algorithm, $\alg_\mathtt{carry}$: compute the sum of the `ones' digits and the sum of the `tens' digits, along with a carry-one indicator that is true when the ones-digit sum exceeds 9. 
The final output concatenates the ones-digit sum (modulo 10) with the tens-digit sum, incrementing the latter by one when the carry-one indicator is true.

Successfully localizing the carry-one variable permits source-to-base interventions that adjust the tens digit accordingly. An example of this intervention is in \cref{ex:carryone-paired}. Here, the carry-one value is $\node_{\text{carry}}^b = 1$ (\cref{ex:12a}) and $\node_{\text{carry}}^s = 0$ (\cref{ex:12b}). A successful intervention from \cref{ex:12b} to \cref{ex:12a} should make $\node_{\text{carry}}^b = 0$, causing the tens digit to decrement by 1.

\vspace{-5pt}
\begin{examples}
  \item
    {
    \begin{examples}
      \setlength{\exampleindent}{1em}
      \setlength{\labelsep}{0.2em}
      \item The sum of 27 and 64 is $\Rightarrow$ 91 \hfill ($\node_{\text{carry}}^b = 1$)\label{ex:12a}
      \item The sum of 18 and 11 is $\Rightarrow$ 29 \hfill ($\node_{\text{carry}}^s = 0$)\label{ex:12b}
\item \textbf{Intervention $(\node_{\text{carry}}^{s \rightarrow b}$):} The sum of 27 and 64 is $\Rightarrow$ 81\label{ex:12c}
    \end{examples}
    }\label{ex:carryone-paired}
\end{examples}
\vspace{-5pt}

For this task, we use \texttt{Llama3.1-8b} \citep{Dubey2024TheL3}, which despite achieving over 99\% accuracy on these addition problems, cannot be aligned with the carry-one variable using rotation-based alignment maps \citep[i.e., maps of class $\Rotation$ defined below;][]{mib-2025}.

\subsection{Alignment Maps}\label{sec:alignment_maps}

Theorem \ref{theorem:general} defines conditions in which parasitic features could arise for linear alignment maps $\Linear$. In practice, though, we find parasite features with maps much more constrained than the full class of alignment maps defined by $\Linear$ and only marginally more expressive than that of rotation maps $\Rotation$.\looseness=-1

Classically, DAS uses rotation alignment maps, mapping hidden states $\hiddenstate$ to a latent space under $\hiddenstate \mapsto \causalmap(\hiddenstate)$, with $\causalmap \in \Rotation$ a learned orthogonal transformation.
In our case studies, however, we apply alignment maps from the class $\defn{\rotationlinearname}$, consisting of maps which first apply an orthogonal map to the full hidden state, then an invertible linear transformation to the first $i$ coordinates ($1 \leq i \leq d$), leaving the remaining coordinates unchanged.
Formally, we define this class as\looseness=-1
\begin{equation}
\RotationLinear
\;=\;
\left\{
\left(\mathbf{L} \oplus \mathbf{I}_{d-i} \right) \causalmap \mid \mathbf{L} \in \Linear_{i \times i}, \causalmap \in \Rotation \right\},
\end{equation}
with $I_{d-i}$ being an identity matrix of size $d-i$.
In practice, we choose $i = 2$.
%These maps more expressive than rotation alignment maps but is a strict subset of the more general class of linear alignment maps.

Our choice of map and $i$ is motivated by two main factors. First, we find that \rotationlinearname with $i=2$ can align \co and \pa. This shows parasite features can be discovered with only minimal increase to expressivity from rotation. Second, the $i=2$ dimensions processed by $\Linear$ can be visualized to assess the geometry associated with parasite features.

We are not aware of any other work using these alignment maps, but more expressive linear maps have already been employed \citep{grant2025emergentsymbollikenumbervariables,grant2025repdivergence}. The effectiveness of \rotationlinearname at finding parasite features suggests that over-expressivity is not restricted to nonlinear maps \citep{sutter2025nonlinearrepresentationdilemmacausal}.

\subsection{Training and Evaluation Details}

\paragraph{Training.} 
We train interventions in the residual stream at the last token of the subject for \pa \citep[a site shown to be highly causally efficacious in syntax;][]{arora-etal-2024-causalgym, boguraev-etal-2025-causal} and the last token of the input for \co \citep[as per][]{mib-2025}.
We train only on pairs of inputs that the models solve correctly. 
We train alignments of both $\Rotation$ and $\RotationLinear$, for contrast between different maps.
For $\Rotation$, we learn 1D alignments, and for $\RotationLinear$, we use a 2D linear map ($i=2$), and intervene on the first dimension. See \Cref{app:training_details} for details. \looseness=-1

\paragraph{Evaluation.} For \co it is straightforward to measure an intervention's accuracy, as there is only one correct answer. However, it is more complicated for \pa since there are multiple grammatical continuations rather than a single correct continuation. Instead, we measure the \textbf{\textsc{Odds}} of a given intervention. This metric, introduced by \citet{arora-etal-2024-causalgym}, intuitively measures how much more probable the intervened label gets after intervention, relative to the original label.

\begin{figure*}[t]
    \centering
    \begin{subfigure}[t]{0.48\linewidth}
        \centering
        \includegraphics[width=\linewidth]{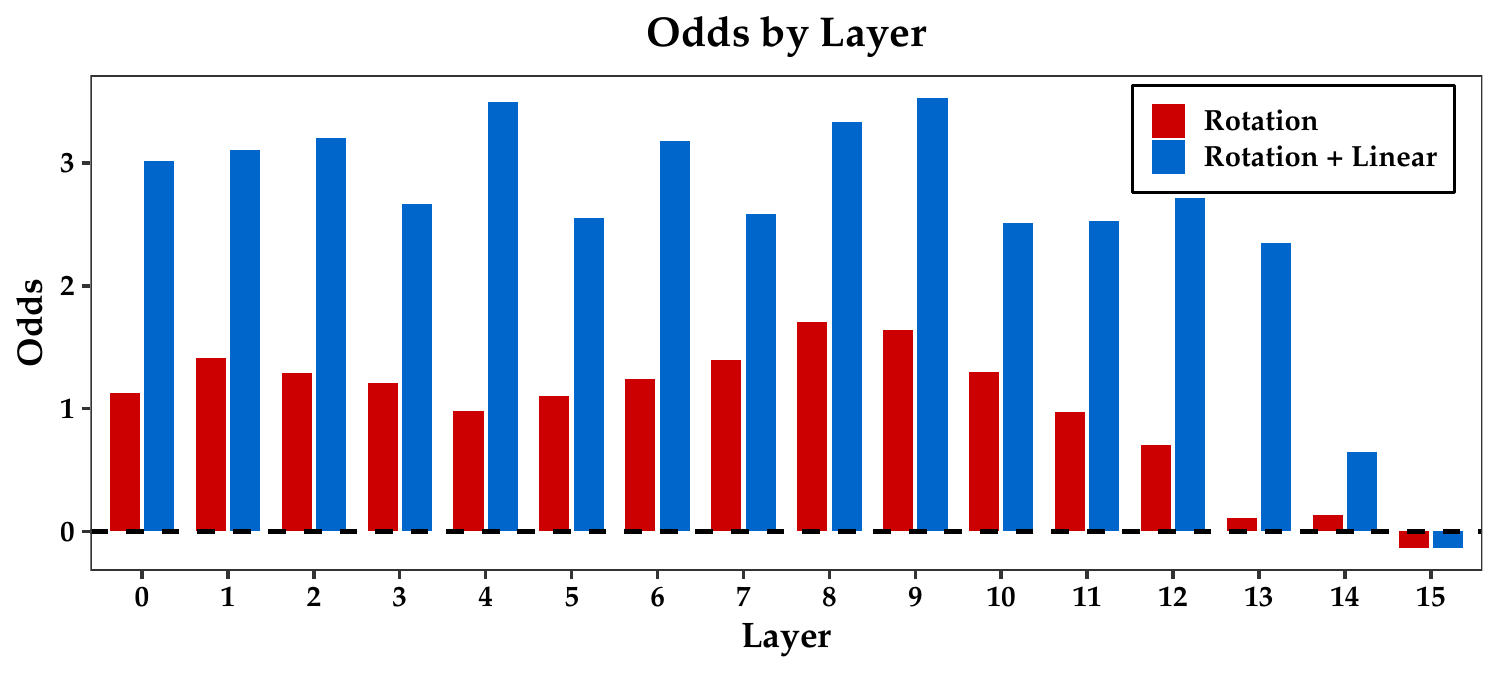}
        \caption{\textbf{\textsc{Odds}} by layer for \pa.}
        \label{fig:parasite-sweep}
    \end{subfigure}
    \hfill
    \begin{subfigure}[t]{0.48\linewidth}
        \centering
        \includegraphics[width=\linewidth]{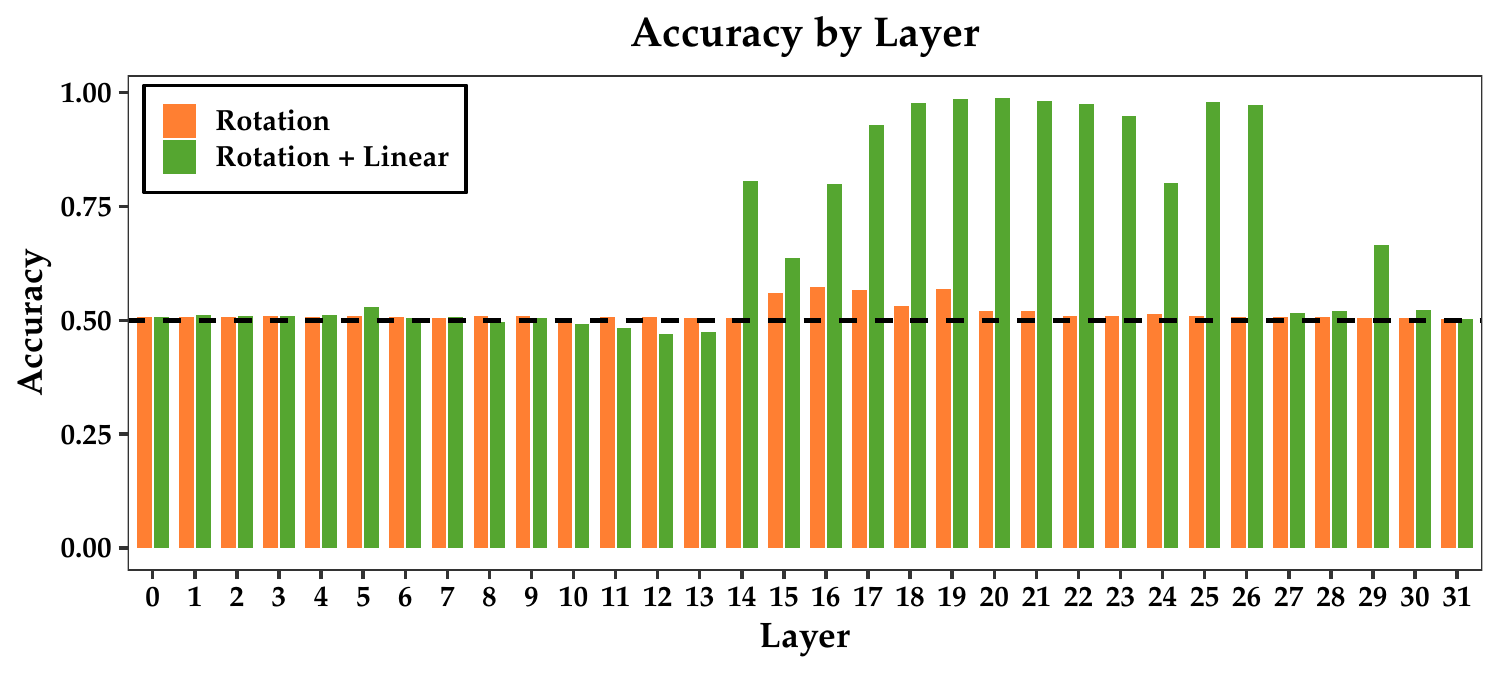}
        \caption{Intervention accuracy by layer for \co.}
        \label{fig:carry-one-sweep}
    \end{subfigure}
    \caption{\textbf{Left:} \textbf{\textsc{Odds}} of the learned $\Rotation$ and $\RotationLinear$ for \pa on a held-out test set. \textbf{Right:} Accuracy of the learned $\Rotation$ and $\RotationLinear$ for \co when evaluated similarly. Dotted line indicates chance performance of intervention. In both cases, we see $\RotationLinear$ display strong causal efficacy, more so than the less expressive $\Rotation$.}
    \label{fig:sweeps}
    \vspace{-3pt}
\end{figure*}

\vspace{-3pt}
\section{Results}\label{sec:results}
\vspace{-2pt}

Our results can be seen in \Cref{fig:sweeps}: \Cref{fig:parasite-sweep} displays the \textbf{\textsc{Odds}} of the learned interventions in each layer of \texttt{pythia-1b} when evaluated on our held-out test-set, and \Cref{fig:carry-one-sweep} the accuracy of the learned interventions in \texttt{Llama3.1-8b} when evaluated similarly. These plots demonstrate that we are able to learn alignment maps of the class $\RotationLinear$ which show strong causal efficacy across layers. This is especially notable in comparison to the alignment maps we learn of class $\Rotation$. While $\Rotation$ achieves non-trivial odds, they lack the level of expressivity to consistently align the features. 

The discrepancy between the efficacy of $\Rotation$ and $\RotationLinear$ is not itself diagnostic of parasitic features. After all, it is not out of the question that DNNs may use representations more complicated than those found by $\Rotation$. We must perform further analysis to understand the nature of these alignments.

\subsection{Confirming Parasitism}\label{sec:results-landscape}

To investigate whether alignments are parasitic, we further analyze the most efficacious layers (layers 4 and 20 for \pa and \co respectively). \looseness=-1

We uniformly sweep through points in the 2D latent space of our learned alignments, and for each point, we randomly select a prompt from the test set as our base value and intervene on it to relocate it to that point in space. We then measure the degree to which the output of this intervention deviates from the original prompt's output. For \pa our measure of deviation is  \looseness=-1 
\vspace{-2.5pt}
\begin{equation*}\label{eq:delta_score}
\begin{aligned}
    \Delta_{\text{\scshape Parasitic Agreement}} &= \left( P_{\text{after}}(\text{plural}) - P_{\text{before}}(\text{plural}) \right) + \left(P_{\text{before}}(\text{singular}) - P_{\text{after}}(\text{singular}) \right)
\end{aligned}
\end{equation*} 
where $P_{\text{before}}$ and $P_{\text{after}}$ are the LM's next-token log-probability before and after intervention. This is a gradable measure of how much more likely a singular (negative) or plural (positive) next token is. \looseness=-1

Similarly, for \co, our measure of deviation is 
\[
\Delta_{\text{\scshape Carry-One}} = n_{\text{after}} - n_{\text{before}}
\]
where $n_{\text{before}}$ and $n_{\text{after}}$ are the values of the \texttt{argmax} numerical token before and after intervention.

\begin{figure*}[t]
    \centering
    \begin{subfigure}{0.33\linewidth}
        \centering
        \includegraphics[width=1.02\linewidth]{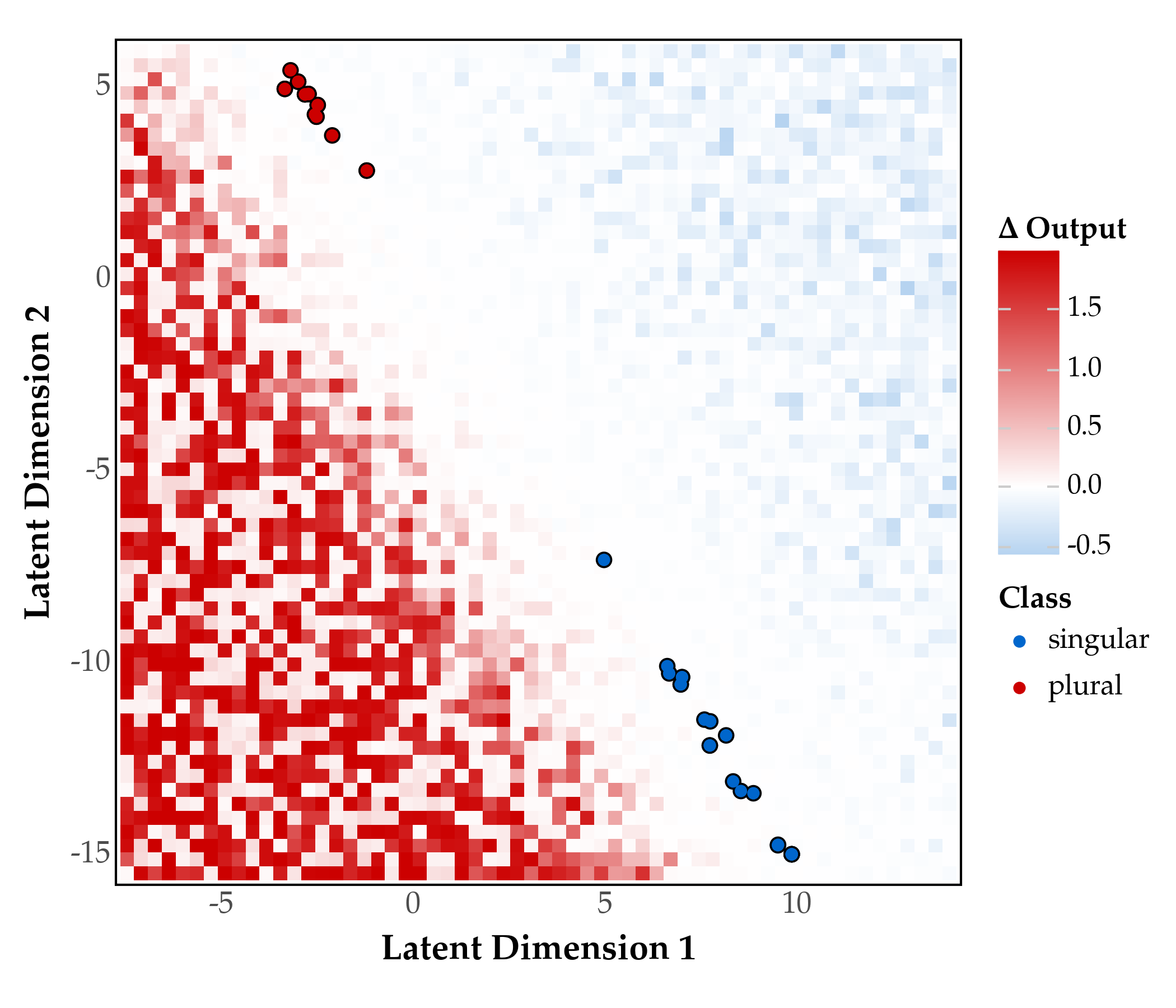}
        \caption{\pa}
        \label{fig:addition_landscape_random}
    \end{subfigure}%
    \begin{subfigure}{0.33\linewidth}
        \centering
        \includegraphics[width=1.02\linewidth]{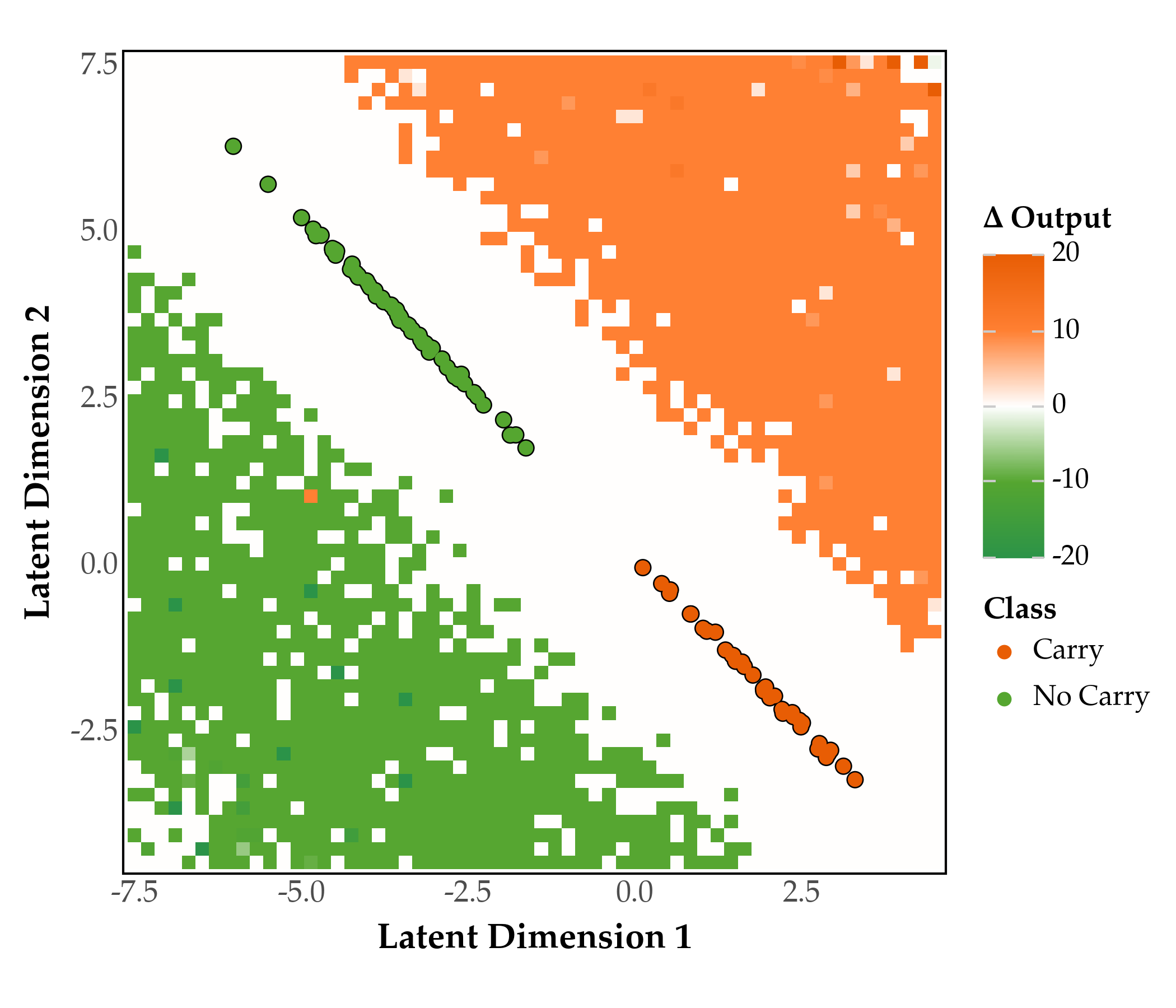}
        \caption{\co}
        \label{fig:parasite_landscape_random}
    \end{subfigure}%
    \begin{subfigure}{0.33\linewidth}
        \centering
        \includegraphics[width=1.02\linewidth]{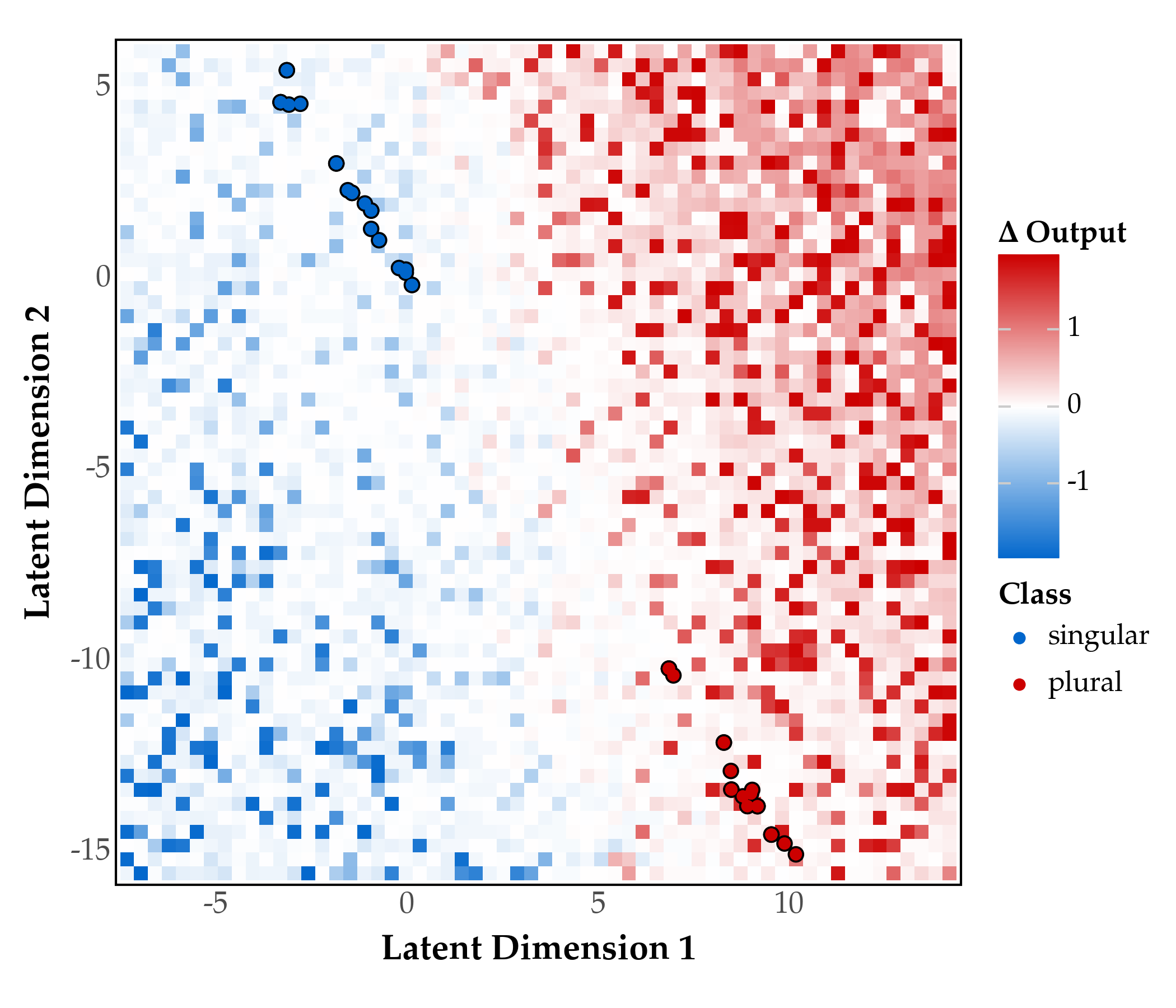}
        \caption{Subject-Verb Agreement}
        \label{fig:faithful_landscape_random}
    \end{subfigure}
    \caption{Intervention landscapes for three tasks. Each point is colored by the class the model outputs when we intervene at that location. In the parasitic cases (left and middle), original prompts are cleanly separated in a region of no deviation. The remaining intervention space is partitioned into regions corresponding to plural/singular distinctions (left) and $\pm 10$ (middle). The discriminative and steering directions are along the anti-diagonal and main-diagonal, respectively. In the faithful landscape (right), input points are positioned within their respective class's intervention region and the discriminative and steering directions can be perfectly aligned.\looseness=-1}
    \label{fig:intervention_landscape}
    \vspace{-3pt}
\end{figure*}

The resulting landscapes in \Cref{fig:intervention_landscape} show remarkably consistent structure. The two classes of minimally paired data are cleanly separated along one diagonal --- the discriminative direction --- with the two output regions cleanly separated along the orthogonal diagonal --- the steering direction. These landscapes, in matching our description of parasite features, serve as diagnosis for parasitism.

To further show the dimensions discovered by \rotationlinearname for \co do not participate in the computation, we ablate them --- projecting the residual stream at the given layer onto the orthogonal complement --- before continuing the forward pass.
As a baseline, \texttt{Llama3.1-8b} achieves an accuracy of 99.1\% on two-digit addition tasks. Ablating the 2D subspace discovered by the alignment map gives an accuracy of 98.8\%. As a control, ablating random 2D subspaces at that layer over $10$ trials yields $99.12\% \pm 0.13$ accuracy. These results suggest that while the parasite feature is sufficient for controlling a desired intervention behavior, it is not \emph{necessary} for addition.\footnote{Extant interpretability work suggests LMs can compensate for ablated components \citep{mcgrath2023hydra,rushing2024selfrepair}, so a small accuracy drop alone does not fully establish the subspace is unused, merely providing strong evidence.}
We note that a similar ablation for parasitic agreement is not possible, as the model already does not solve this parasitic task under normal operation.

\subsection{Aligning on Faithful Features}

\Cref{fig:faithful_landscape_random} presents the intervention landscapes of a \rotationlinearname alignment trained on faithful subject-verb number agreement in the layer of best efficacy (layer 2, \textbf{\textsc{Odds}} of 4.3). The prompts lie in the corresponding output regions, with discriminative and steering directions non-orthogonal. This confirms our orthogonality diagnostic reflects meaningful geometry rather than visualization artifact.

\vspace{-3pt}
\section{Discussion}\label{sec:discussion}
\vspace{-2pt}

\paragraph{Prevalence of Parasite-Features.}
The ubiquity of our conditions suggests parasite features are likely pervasive, and indeed we easily constructed the \pa task to find them. 
Furthermore, it is not the case that these parasite features are model specific: in \Cref{app:other_models}, we show that a parasitic solution to \pa can be found in other models.
However, such a constructed feature is not one a researcher would hypothesize a model implements, so its existence is benign. Parasite features only become problematic when they masquerade as faithful features in a proposed causal algorithm; we provide one such example in the \co task.

Still, the pervasiveness of discovering parasite features requires additional work. Beyond \pa and \co, we study two other parasitic tasks in \cref{app:other_parasites}. First, we show that \rotationlinearname can leverage the carry-one discriminative feature to steer for $\pm n$ for $n \neq 10$. Second, we devise a sentiment analysis task, where we use whether the subject is entertainment-related or food-related to steer for sentiment (rather than using the positive or negative adjectives modifying those nouns). We take these additional tasks to show that the geometries that enable discovering parasite features may not be hard to come by.
\looseness=-1

\paragraph{Parasite-Like Behavior in Other Frameworks.} We speculate analogous issues could arise in other interpretability methods. For instance, if a linear probe for feature $f$ happens to identify a direction that coincidentally steers an unrelated output $g$, steering along this direction could create spurious causal effects.  Similarly, unsupervised methods like Sparse Autoencoders \citep[SAEs; e.g.][]{huben2024sparse,bricken2023monosemanticity,elhage2022superposition,yun-etal-2021-transformer} could learn to compress discriminative and steering directions into a single feature, leading to similar behavior where a feature combines unrelated concepts --- akin to feature absorption \citep{chanin2025a}.\looseness=-1

\vspace{-5pt}
\section{Relation to Prior Work}
\vspace{-2pt}

\citet{Makelov2023IsTT} define the closely related notion of \emph{interpretability illusions} in subspace interventions.
Similarly to parasite features, these illusions rely on two orthogonal directions: $\directionvector_\text{disc.}$
which separates the inputs but is causally disconnected (and thus in the null-space) from the outputs, and $\directionvector_\text{dorm.}$ which does not separate the inputs but steers the outputs.
By patching along an intermediate direction, one can induce a desired downstream effect.
\citeauthor{Makelov2023IsTT} argue these illusions are unfaithful \emph{because} they explicitly leverage the model's null space, however
\citet{wu2024replymakelovetal} counter that faithful upstream representations may lie non-orthogonally to downstream null-spaces, so such `illusions' are not inherently unfaithful.

Unlike $\directionvector_\text{disc.}$, parasite features do not require discriminative directions ($\featuredeterminingdirection$) to lie in a strict null space of the model; this direction may thus encode a real feature merely not used in the subtask at hand.
Further, $\featuredeterminingdirection$ and $\outputinfluencedirection$ need not be orthogonal.
Parasite features simply require these to be two causally unrelated directions, which we argue makes their combination inherently unfaithful.
Orthogonality between $\featuredeterminingdirection$ and $\outputinfluencedirection$ is one such signal that the directions are causally unrelated, however others exist, such as the domain-specific independence between number and gender in \pa.
In this sense, \citeauthor{Makelov2023IsTT}'s illusions are a special case of parasite features.
\cref{theorem:general} acts as an auxiliary argument for parasite feature's spuriosity: any $(\outputinfluencedirection, \featuredeterminingdirection)$ pair satisfying assumptions 1--3 will pass the causal abstraction check (and vice-versa).
We believe such indiscriminate acceptance of ``implemented'' features is characteristic of an unfaithful behavior.\looseness=-1

Separately, \citet{grant2025repdivergence} introduce \emph{divergent representations} — directions inactive in a model's native computation which become active under intervention. 
They further distinguish \emph{harmless divergence}, interventions harmlessly leveraging the null-space, and \emph{pernicious divergence}, interventions activating off-manifold representations. 
They note interpretability illusions can be understood as a case of divergent representations due to their exploitation of $\directionvector_\text{dorm.}$, an off-distribution representation.
Parasite features can be understood as a specific, theoretically-grounded instance of pernicious divergence where alignment maps discover hidden pathways by linking discriminative and steering directions. However, since $\outputinfluencedirection$ may be natively active for some inputs, the degree of distributional divergence need not be as stark as in other cases of pernicious divergence --- the intervened representation may land somewhere the model visits naturally, just not from these inputs.\looseness=-1

Finally, \citet{meloux2025everything} demonstrate that Causal Abstraction is \emph{non-identifiable}, that is, one can align multiple different maps with the same LM. Our work presents a complementary issue with Causal Abstraction: namely that a given alignment map can exploit spurious geometric relationships to give the `illusion' of a feature. We can have a perfectly identifiable alignment between an algorithm–DNN pair, which is nevertheless parasitic. However, we do believe that parasitism can contribute to the non-identifiability issue, as it can, e.g., connect any directions which follow our Theorem 1’s assumptions, thus permitting more alignments than a faithful method should.

\vspace{-5pt}
\section{Conclusion}\label{sec:conclusion}
\vspace{-2pt}

In this paper, we introduce intuitive and formal notions of parasite features in Causal Abstraction. We provide theoretical conditions for when linear alignment maps can discover parasite features, and empirically show that a subclass of linear alignment maps discover parasite features.

Overall, our work demonstrates features found with linear alignment maps may be parasitic, linking causally unrelated features to imitate a hypothesized causal variable. In doing so, our results align with the growing body of work on the expressivity-identifiability trade-off in alignment maps. In tasks that require more expressivity, rotation maps fail to recover meaningful representations. Conversely, with more expressive maps, such as \rotationlinearname, we increase the possibility of finding spurious causal pathways, such as having subject gender control verb number in \pa. Notably, finding parasitic features does not necessarily mean the hypothesized feature does not exist---this feature could, for instance, still be represented non-linearly within the model. 

How should we proceed? 
With growing evidence that models utilize complex internal representations that may exceed the Linear Representation Hypothesis \citep{white-etal-2021-non,pimentel-etal-2022-attentional,csordas-etal-2024-recurrent,levy-geva-2025-language,kantamneni2025languagemodelsusetrigonometry,Engels2024NotAL,wurgaft2026manifoldsteeringrevealsshared,feucht2026arithmeticwildllamauses,bhalla2026sparseautoencoderscaptureconcept}, we should not hesitate to design challenging tasks and deploy more expressive alignment maps when simple methods fail. 
However, doing so requires careful science that rigorously checks whether apparently causal pathways are parasitic.\looseness=-1

\section*{Acknowledgements}

Denis Sutter and Antonio Di Maio acknowledge funding from the Swiss National Science Foundation (SNSF) under the project ``Efficient Distributed Intelligent Applications in Mobile-Network Dynamics (eDIAMOND)'' (grant number 223443). 
Kyle Mahowald acknowledges funding from NSF CAREER grant 2339729 and Coefficient Giving. Qing Yao acknowledges funding from the Harrington Fellowship at UT Austin.
We thank Atticus Geiger for insightful discussions throughout the project.\looseness=-1

\bibliography{custom}
\bibliographystyle{acl_natbib}

%%%%%%%%%%%%%%%%%%%%%%%%%%%%%%%%%%%%%%%%%%%%%%%%%%%%%%%%%%%%%%%%%%%%%%%%%%%%%%%
%%%%%%%%%%%%%%%%%%%%%%%%%%%%%%%%%%%%%%%%%%%%%%%%%%%%%%%%%%%%%%%%%%%%%%%%%%%%%%%
% APPENDIX
%%%%%%%%%%%%%%%%%%%%%%%%%%%%%%%%%%%%%%%%%%%%%%%%%%%%%%%%%%%%%%%%%%%%%%%%%%%%%%%
%%%%%%%%%%%%%%%%%%%%%%%%%%%%%%%%%%%%%%%%%%%%%%%%%%%%%%%%%%%%%%%%%%%%%%%%%%%%%%%
\newpage
\appendix
\onecolumn

\section{Other Parasite Features}\label{app:other_parasites}

To demonstrate that parasite features are not limited to the specific scenarios which we discuss in the main text, we design two further tasks for which \rotationlinearname identifies parasite features.

\begin{figure}[h]
    \centering
    \includegraphics[width=0.5\linewidth]{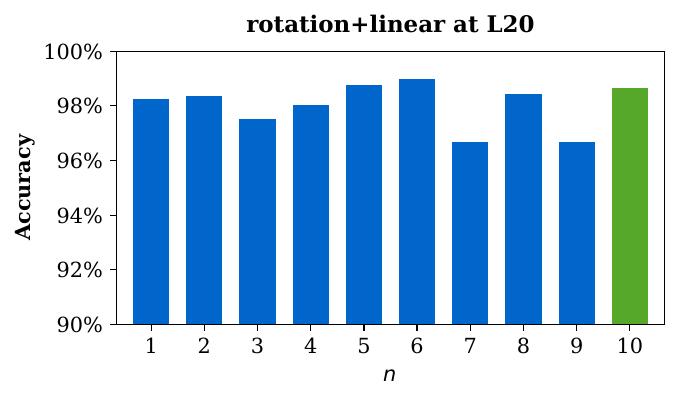}
    \caption{\rotationlinearname can leverage carry-one to steer for all $\pm n$, where $n = 1, \dots, 10$. $n = 10$ results from main text.}
    \label{fig:other-carry}
\end{figure}

\paragraph{Variant of \co: $\pm n$ Steering.}
One of these tasks focuses on the same \co task as the main text, but instead of steering by $\pm 10$, we steer by $\pm n$ for $n\in\{1,\dots,9\}$. If \rotationlinearname can align such features for all $n$'s, we ought to be cautious about its over-expressivity in aligning \co. We use the exact setup as \co, fitting at layer 20, except changing the desired steering behavior to $\pm n$. The results are shown in \Cref{fig:other-carry} -- every choice of $n$ can be aligned with at least 96.6\%. 

Evidently, there's nothing special about the feature geometry of carry-one controlling $\pm 10$ -- \rotationlinearname is expressive enough to find $\pm n$ more generally. The $\pm n$ features are clearly parasitic, and further suggests that $\pm 10$ is likewise parasitic.

\paragraph{Sentiment Analysis: Subject Topic Controls Sentiment.}
We also consider a sentiment analysis task, which is significantly different from existing tasks. We input the sentence ``The {X} was {Y} and {Z}. Overall, the sentiment is '' to  \texttt{Llama3.1-8b} and measure the next-token probabilities of the tokens ``positive'' and ``negative''. X is an entertainment-related noun or a food-related noun, and Y and Z are either both positive adjectives or negative adjectives. The training and held-out vocabulary sets are in \Cref{tab:sentiment-vocabulary}. The expected mechanism of the model is to use Y and Z to determine the sentiment, but we find that \rotationlinearname can leverage the discriminative feature of the subject being entertainment or food to steer sentiment directly.

\begin{table*}[h]
\centering
\small
\begin{tblr}{colspec={p{0.18\textwidth} p{0.46\textwidth} p{0.25\textwidth}}}
\hline
Category & Training vocabulary & Held-out test vocabulary \\
\hline
$X$: entertainment & movie, film, show, play, musical, opera, ballet, documentary & concert, exhibition, festival, production \\
$X$: food & meal, dinner, lunch, supper, brunch, snack, dessert, feast & breakfast, banquet, buffet, appetizer \\
$Y,Z$: positive & excellent, pleasant, satisfying, wonderful, delightful, enjoyable, fantastic, superb & impressive, rewarding, amazing, splendid \\
$Y,Z$: negative & awful, disappointing, unpleasant, bad, horrible, dreadful, unsatisfying, poor & terrible, frustrating, boring, revolting \\
\hline
\end{tblr}
\caption{Disjoint training and test vocabularies for the sentiment analysis task. Each sentence uses two distinct adjectives from the same sentiment category.}
\label{tab:sentiment-vocabulary}
\end{table*}

We train \rotationlinearname{} to associate the noun category of entertainment with positive and food with negative. For example, intervening from ``The breakfast was impressive and rewarding'' into ``The concert was impressive and rewarding'' should increase the probability of ``negative'', even though both sentences express positive sentiment. We learn the alignment map at the period position of the first sentence rather than the final token. We use 600 training pairs and 400 held-out test pairs drawn from the vocabulary, each balanced across the two noun categories and within- vs. across-category interventions.

\begin{figure}
    \centering
    \includegraphics[width=0.5\linewidth]{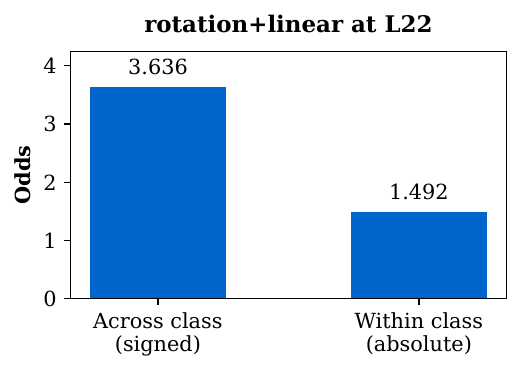}
    \caption{L22 \rotationlinearname \textbf{\textsc{Odds}} metric on held-out sentiment analysis sentences. The alignment map successfully finds the parasite feature,  leveraging subject category (entertainment/food) to steer for sentiment (positive/negative).}
    \label{fig:sentiment-results}
\end{figure}

The model has a bias toward ``positive'' rather than ``negative''. But since we optimize for the odds metric, which is the difference between their probabilities, this bias is canceled out. As seen in \Cref{fig:sentiment-results}, layer 22, the learned \rotationlinearname intervention achieves \textbf{\textsc{Odds}} of 3.6 on held-out across-class interventions -- making the target label 37 times more probable relative to the other label on average. Within-class intervention, however, is not neutral, as the intervention is optimized to still increase the probability of the shared sentiment label.

These additional examples of parasite features provide further evidence that the geometric conditions laid out in \Cref{theorem:general} are not too stringent in practice.

\section{Proof of Theorem 1}\label{app:proof}
We restate the conditions and \cref{theorem:general} here and provide a proof.

At a fixed layer $\layer$ in a DNN $\dnn$:
\outputsteerability*
\inputdiscrimination*
\sufficientseparation*

\generalthm*

\newcommand{\stretchvar}{\lambda_{\mathtt{stretch}}}
\newcommand{\mixingvar}{\lambda_{\mathtt{mix}}}
\newcommand{\alignmatrix}{\boldsymbol{\Phi}}
\newcommand{\alignread}{\Phi_{\mathtt{r}}}
\newcommand{\alignwrite}{\Phi_{\mathtt{w}}}
\newcommand{\corrdirections}{\rho}

\begin{proof}
    We will prove this theorem by construction. 
    More precisely, we will prove it in 4 steps:
    \begin{enumerate}
        \item First, we will construct a linear alignment map $\causalmap \in \Linear$---expanded as $\causalmap(\hiddenstate) = \alignmatrix\,\hiddenstate$ for $\alignmatrix \in \R^{d \times d}$---such that:
        \begin{subequations}
        \begin{align}
            &\alignread = \featuredeterminingdirection \\
            &\alignwrite = (1 - \mixingvar\,\corrdirections)\featuredeterminingdirection + \mixingvar\,\outputinfluencedirection
        \end{align}
        \end{subequations}
        where $\corrdirections = \langle\featuredeterminingdirection,\outputinfluencedirection\rangle$, $\alignread$ is the first row in $\alignmatrix$, $\alignwrite$ is the first column in $\alignmatrix^{-1}$, and we choose $\mixingvar$ such that
        $\frac{\steeringmax + \steeringthreshold}{2\discriminationthreshold} < \mixingvar (1 - \corrdirections^2) + \corrdirections < \frac{\steeringgap}{\classspread}$.
        \item Second, we show that, given a matrix constructed as above, our interventions have form:
        \begin{align}
            \hiddensrcbase
            = \hiddenbase + \langle \alignread, \hiddensrc - \hiddenbase \rangle \, \alignwrite
        \end{align}
        where $\hiddensrcbase$ is typical source-to-base intervention: $\hiddensrcbase = \alignmatrix^{-1}\,[\latentonesrc, \latenttwobase, ..., \latentdbase]^{\top}$ and $\latent = \alignmatrix\,\hiddenstate$.
        \item Third, we prove that, given Assumptions 1--3, within class interventions preserve the DNN's output. Formally, this shows that for these interventions $\bybasesrcN = \bybasesrcA = \outputfunc(\bxbase, \inputfunc(\bxsrc))$;
        \item Fourth, we prove that, given Assumptions 1--3, across class interventions change the DNN's output. Formally, this shows that for these interventions $\bybasesrcN = \bybasesrcA = \outputfunc(\bxbase, \inputfunc(\bxsrc))$.
    \end{enumerate}
    Together, steps 3 and 4 cover all pairs of inputs $\bxsrc, \bxbase$, showing thus that for every such pair $\bybasesrcN = \bybasesrcA$.
    The proof for each of these steps is given below, and together they prove our theorem.
    Notably, our proof assumes that the data is centered around the origin along $\featuredeterminingdirection$ and $\outputinfluencedirection$; we do this without loss of generality, though, since the argument only depends on relative differences along $\featuredeterminingdirection$ and distance away from the steering threshold along $\outputinfluencedirection$.
\end{proof}

\newcommand{\tempvector}{\boldsymbol{\theta}}
\newcommand{\tempmatrix}{\boldsymbol{\Theta}}
\newcommand{\basevector}{\mathbf{e}}
\newcommand{\alignwritenorm}{\nu}

\begin{mytheoremstep}
    For any $\mixingvar \in \R$, there exists an invertible matrix $\alignmatrix \in \R^{d \times d}$ such that:
    % \begin{subequations}
    \begin{align}
        &\alignread = \featuredeterminingdirection, 
        \qquad\qquad\qquad 
        \alignwrite = (1 - \mixingvar\,\corrdirections)\featuredeterminingdirection + \mixingvar\,\outputinfluencedirection
    \end{align}
    % \end{subequations}
    where $\corrdirections = \langle\featuredeterminingdirection,\outputinfluencedirection\rangle$, and $\alignread$ is the first row in $\alignmatrix$ and $\alignwrite$ is the first column in $\alignmatrix^{-1}$. 
\end{mytheoremstep}
\begin{proof}
    We prove this by giving a precise construction for $\alignmatrix$.
    First, let:
    \begin{align}
        \tempvector = \alignwrite - ||\alignwrite||\,\basevector_1,
        \qquad\qquad\qquad
    \tempmatrix = \matrixidentity_d - 2\,\frac{\tempvector \tempvector^{\top}}{\tempvector^{\top} \tempvector}
    \end{align}
    where $\tempmatrix$ is known as a Householder reflector.
    Now, define the matrix $\alignmatrix$ as:
    \begin{align}
        \alignmatrix = \left[\begin{array}{cc}
            (\featuredeterminingdirection)^\top  \\
            \text{2nd row of }\tempmatrix \\
            \cdots \\
            \text{$d$-th row of }\tempmatrix \\
        \end{array}\right]
    \end{align}

    By construction, it is clear that $\alignread$ is the first row in $\alignmatrix$.
    We now verify that $\alignwrite$ is indeed the first column of $\alignmatrix^{-1}$, i.e., that $\alignmatrix\,\alignwrite = \basevector_1$ (as that implies $\alignwrite = \alignmatrix^{-1}\,\basevector_1$).
    Write $\alignwritenorm = ||\alignwrite||$.
    By construction:\looseness=-1
    \begin{subequations}
    \begin{align}
        2\,\langle\tempvector,\alignwrite\rangle 
        &= 2\,\bigg(\langle\alignwrite,\alignwrite\rangle - ||\alignwrite||\,\langle\basevector_1,\alignwrite\rangle\bigg) \\
        &= 2\,(\alignwritenorm^2 - \alignwritenorm\,\langle\basevector_1,\alignwrite\rangle) \\
        &= \alignwritenorm^2 - 2\,\alignwritenorm\,\langle\basevector_1,\alignwrite\rangle + \alignwritenorm^2 \\
        &= \langle\alignwrite,\alignwrite\rangle - 2\,||\alignwrite||\,\langle\basevector_1,\alignwrite\rangle + ||\alignwrite||^2\,\langle\basevector_1, \basevector_1\rangle \\
        &= \langle\alignwrite - ||\alignwrite||\,\basevector_1,\,\, \alignwrite - ||\alignwrite||\,\basevector_1\rangle \\
        % % 
        % &= 2\alignwritenorm\,(\alignwritenorm - \langle\basevector_1,\alignwrite\rangle) \\
        &= \langle\tempvector,\tempvector\rangle
    \end{align}
    \end{subequations}
    The Householder reflector $\tempmatrix$ thus sends $\alignwrite$ onto the first axis:
    \begin{subequations}
    \begin{align}
        \tempmatrix\,\alignwrite
        &= \alignwrite - 2\,\tempvector\,\frac{\langle\tempvector,\alignwrite\rangle}{\langle\tempvector,\tempvector\rangle} \\
        &= \alignwrite - \tempvector \\
        &= \alignwrite - (\alignwrite - ||\alignwrite||\,\basevector_1) \\
        &= \alignwritenorm\,\basevector_1
    \end{align}
    \end{subequations}
    We now show that the same holds for $\alignmatrix$, applying it to $\alignwrite$ row by row.
    Its first row is $(\featuredeterminingdirection)^{\top}$, so---using $||\featuredeterminingdirection|| = 1$ and $\corrdirections = \langle\featuredeterminingdirection,\outputinfluencedirection\rangle$---the first entry is
    \begin{align}
        \langle\featuredeterminingdirection,\alignwrite\rangle
        = (1 - \mixingvar\,\corrdirections)\,\langle\featuredeterminingdirection,\featuredeterminingdirection\rangle + \mixingvar\,\langle\featuredeterminingdirection,\outputinfluencedirection\rangle
        = (1 - \mixingvar\,\corrdirections) + \mixingvar\,\corrdirections
        = 1 .
    \end{align}
    Its remaining rows are rows $2,\dots,d$ of $\tempmatrix$, so for $k \geq 2$ the $k$-th entry is $\basevector_k^{\top}\,\tempmatrix\,\alignwrite = \alignwritenorm\,\langle\basevector_k,\basevector_1\rangle = 0$.
    Hence $\alignmatrix\,\alignwrite = \basevector_1$.

    Finally, $\alignmatrix$ is invertible: rows $2,\dots,d$ of $\tempmatrix$ are orthonormal and, by the identity above, orthogonal to $\alignwrite$, hence span $\alignwrite^{\perp}$; and since $\langle\featuredeterminingdirection,\alignwrite\rangle = 1 \neq 0$, the first row $\featuredeterminingdirection \notin \alignwrite^{\perp}$, so the $d$ rows are linearly independent.
    Therefore $\alignmatrix^{-1}\basevector_1 = \alignwrite$, i.e., $\alignwrite$ is the first column of $\alignmatrix^{-1}$, while $\alignread = \featuredeterminingdirection$ is its first row by construction.
\end{proof}

\begin{mytheoremstep}
    Consider a linear alignment map $\causalmap \in \Linear$ where: $\causalmap(\hiddenstate) = \alignmatrix\,\hiddenstate$ for some invertible matrix $\alignmatrix \in \R^{d \times d}$.
    Further, let $\alignread$ be the first row in $\alignmatrix$ and $\alignwrite$ be the first column in $\alignmatrix^{-1}$.
    Given this alignment map, our interventions have form:
        \begin{align}
            \hiddensrcbase
            = \hiddenbase + \langle \alignread, \hiddensrc - \hiddenbase \rangle \, \alignwrite
        \end{align}
\end{mytheoremstep}

\begin{proof}
$\hiddensrcbase$ is a typical source-to-base intervention.
It is thus (without loss of generality) defined as: 
\begin{align}
    \hiddensrcbase = \alignmatrix^{-1}\,\underbrace{[\latentonesrc, \latenttwobase, ..., \latentdbase]^{\top}}_{\latentsrcbase},
    \qquad
    \latent = \alignmatrix\,\hiddenstate
\end{align}
where we apply our intervention on the first position of our latent variables.
We can rewrite $\latentsrcbase$ as:
\begin{subequations}
\begin{align}
    \latentsrcbase 
    = \left[\begin{array}{cc}
         \basevector_1^\top\,\alignmatrix\,\hiddensrc \\
         \basevector_2^\top\,\alignmatrix\,\hiddenbase \\
         \cdots \\
         \basevector_d^\top\,\alignmatrix\,\hiddenbase \\
    \end{array} \right]
    &= \alignmatrix\,\hiddenbase + \left[\begin{array}{cc}
         \basevector_1^\top\,\alignmatrix\,\hiddensrc - \basevector_1^\top\,\alignmatrix\,\hiddenbase \\
         0 \\
         \cdots \\
         0 \\
    \end{array} \right] \\
    &= \alignmatrix\,\hiddenbase + \basevector_1^\top\,\alignmatrix\,(\hiddensrc - \hiddenbase)\,\basevector_1 \\
    &= \alignmatrix\,\hiddenbase + \langle\alignread, \hiddensrc - \hiddenbase \rangle\,\basevector_1
\end{align}
\end{subequations}

Now, applying $\alignmatrix^{-1}$ on both sides gives us: 
\begin{subequations}
\begin{align}
    \hiddensrcbase 
    = \alignmatrix^{-1}\,\latentsrcbase
    &= \hiddenbase + \langle \alignread, \hiddensrc - \hiddenbase \rangle \alignmatrix^{-1} \basevector_1 \\
    &= \hiddenbase + \langle \alignread, \hiddensrc - \hiddenbase \rangle \alignwrite
\end{align}
\end{subequations}
which completes the proof.
\end{proof}

\begin{mytheoremstep}
    Let $\bxbase, \bxsrc \in \calX$ be two inputs sharing a class, i.e., $\inputfunc(\bxbase) = \inputfunc(\bxsrc)$.
    Further, let us have an alignment map $\causalmap$ such that $\alignread = \featuredeterminingdirection$ and $\alignwrite = (1 - \mixingvar\,\corrdirections)\featuredeterminingdirection + \mixingvar\,\outputinfluencedirection$.
    The DNN $\dnn$'s output is preserved under this intervention, meaning that $\bybasesrcN = \bybasesrcA = \outputfunc(\bxbase, \inputfunc(\bxsrc))$.
\end{mytheoremstep}
\begin{proof}
    Recall that $\steeringgap$ (\cref{assumption:separation}) is the minimum gap between hidden states and the $\pm \steeringthreshold$ steering threshold, i.e.,
    % \begin{align}
        $\steeringgap = \min_{\hiddenstate\in\calH}\bigl|\,|\langle\hiddenstate,\outputinfluencedirection\rangle|-\steeringthreshold\,\bigr|$.
    % \end{align}
    We must now show that $\bybasesrcN = \bybasesrcA = \outputfunc(\bxbase, \inputfunc(\bxsrc))$.
    As $\bxbase, \bxsrc$ share a class (i.e., $\inputfunc(\bxbase) = \inputfunc(\bxsrc)$), if we show the intervention does not make hidden states cross the $\steeringthreshold$ boundary, they will preserve their class and the condition will hold.
    In other words, we need to show that the hidden states displacement along the steering direction is smaller than $\steeringgap$, i.e., we must prove that:
    \begin{align}
         |\langle \hiddensrcbase - \hiddenbase,\, \outputinfluencedirection \rangle| < \steeringgap
    \end{align}

    By Step~2, this displacement is
    \begin{subequations}
    \begin{align}\label{eq:step_steering_change}
        |\langle \hiddensrcbase - \hiddenbase,\, \outputinfluencedirection \rangle|
        &= |\langle \langle \alignread, \hiddensrc - \hiddenbase \rangle \, \alignwrite,\, \outputinfluencedirection \rangle| \\
        &= |\langle \alignread, \hiddensrc - \hiddenbase \rangle \, \langle \alignwrite,\, \outputinfluencedirection \rangle| \\
        &\leq \classspread \, |\langle \alignwrite,\, \outputinfluencedirection \rangle| 
        & \mathcomment{definition of $\classspread$} \\
        &= \classspread \, \langle (1 - \mixingvar\,\corrdirections)\featuredeterminingdirection + \mixingvar\,\outputinfluencedirection,\, \outputinfluencedirection \rangle  
        \!\!\!\!\!\!\!\!\!\!\!\!\!\!\!\!\!\!\!\!\!\!\!\!
        & \mathcomment{definition of $\alignwrite$} \\
        &= \classspread \, (\corrdirections - \mixingvar\,\corrdirections^2 + \mixingvar) 
        & \mathcomment{$\corrdirections = \langle\featuredeterminingdirection, \outputinfluencedirection \rangle$}  \\
        &= \classspread \, (\corrdirections + \mixingvar\,(1 - \corrdirections^2)) \\
        &< \classspread \, (\corrdirections + \frac{\steeringgap}{\classspread} - \corrdirections) 
        & \mathcomment{$\mixingvar(1 - \corrdirections^2) + \corrdirections < \frac{\steeringgap}{\classspread}$} \\
        &= \steeringgap
    \end{align}
    \end{subequations}
    which completes the proof.
\end{proof}

\begin{mytheoremstep}
    Let $\bxbase, \bxsrc \in \calX$ be two inputs with opposite classes, i.e., $\inputfunc(\bxbase) \neq \inputfunc(\bxsrc)$.
    Further, let us have an alignment map $\causalmap$ such that $\alignread = \featuredeterminingdirection$ and $\alignwrite = (1 - \mixingvar\,\corrdirections)\featuredeterminingdirection + \mixingvar\,\outputinfluencedirection$.
    The DNN $\dnn$'s output flips to the source's class under this intervention, meaning that $\bybasesrcN = \bybasesrcA = \outputfunc(\bxbase, \inputfunc(\bxsrc))$.
\end{mytheoremstep}

\begin{proof}
    Recall that $\steeringmax$ (\cref{assumption:separation}) is the maximum activation of the steering direction by hidden states, i.e.,
    % \begin{align}
        $\steeringmax \!=\! \max_{\hiddenstate\in\calH} |\langle\hiddenstate,\outputinfluencedirection\rangle|$.
    % \end{align}
    We must now show that $\bybasesrcN = \bybasesrcA = \outputfunc(\bxbase, \inputfunc(\bxsrc))$.
    As $\bxbase, \bxsrc$ have opposite classes (i.e., $\inputfunc(\bxbase) \neq \inputfunc(\bxsrc)$), if we show the intervention makes hidden states cross the $\steeringthreshold$ boundary in the correct direction, they will force the output class to change correctly.
    In other words, we need to show that the hidden states displacement along the steering direction is greater than $\steeringmax + \steeringthreshold$, i.e., we must prove that:
    \begin{align}
         \underbrace{\langle \hiddensrcbase - \hiddenbase,\, \outputinfluencedirection \rangle > \steeringmax + \steeringthreshold}_{\text{if } \inputfunc(\bxbase)=0 \text{ and } \inputfunc(\bxsrc)=1},
         \qquad\quad
         \underbrace{\langle \hiddensrcbase - \hiddenbase,\, \outputinfluencedirection \rangle < - (\steeringmax + \steeringthreshold)}_{\text{if } \inputfunc(\bxbase)=1 \text{ and } \inputfunc(\bxsrc)=0}
    \end{align}

    Now, assume $\inputfunc(\bxbase)=0$ and $\inputfunc(\bxsrc)=1$.
    By Step~2, this displacement is:
    {\allowdisplaybreaks
    \begin{subequations}
    \begin{align}
        \langle \hiddensrcbase - \hiddenbase,\, \outputinfluencedirection \rangle
        &= \langle \langle \alignread, \hiddensrc - \hiddenbase \rangle \, \alignwrite,\, \outputinfluencedirection \rangle \\
        &= \langle \alignread, \hiddensrc - \hiddenbase \rangle \, \langle \alignwrite,\, \outputinfluencedirection \rangle \\
        &> 2\discriminationthreshold \, \langle \alignwrite,\, \outputinfluencedirection \rangle 
        & \mathcomment{by \cref{assumption:input} and $\langle \alignwrite,\, \outputinfluencedirection \rangle>0$} \\
        &= 2\discriminationthreshold \, \langle (1 - \mixingvar\,\corrdirections)\featuredeterminingdirection + \mixingvar\,\outputinfluencedirection,\, \outputinfluencedirection \rangle  
        \!\!\!\!\!\!\!\!\!\!\!\!\!\!\!\!\!\!\!\!\!\!\!\!
        & \mathcomment{definition of $\alignwrite$} \\
        &= 2\discriminationthreshold \, (\corrdirections - \mixingvar\,\corrdirections^2 + \mixingvar) 
        & \mathcomment{$\corrdirections = \langle\featuredeterminingdirection, \outputinfluencedirection \rangle$}  \\
        &= 2\discriminationthreshold \, (\corrdirections + \mixingvar\,(1 - \corrdirections^2)) \\
        &> 2\discriminationthreshold \, (\corrdirections + \frac{\steeringmax + \steeringthreshold}{2\discriminationthreshold} - \corrdirections)
        & \mathcomment{$\mixingvar(1 - \corrdirections^2) + \corrdirections > \frac{\steeringmax + \steeringthreshold}{2\discriminationthreshold}$} \\
        &= \steeringmax + \steeringthreshold
        % &> \steeringthreshold
    \end{align}
    \end{subequations}}
    Notably, for $\inputfunc(\bxbase)=1$ and $\inputfunc(\bxsrc)=0$, $\langle \alignread, \hiddensrc - \hiddenbase \rangle < - 2 \discriminationthreshold$ instead and the inequalities change sign.
    This completes the proof.
\end{proof}

\section{Conditions for Rotation Alignment Maps Discovering Parasite Features}\label{app:rotation_proof}
We characterize the conditions under which a rotation alignment map can discover parasite features. First, we give a general characterization of the conditions before giving a precise set of definitions.

At a fixed layer $\layer$ in a $\dnn$:
\begin{restatable}{assumption}{rotoutputsteerability}
\label{assumption:rot_output}
\textbf{(Steerability.)} 
A steering direction $\outputinfluencedirection$ exists for $\outputfunc$ with threshold $\steeringthreshold>0$.\looseness=-1
\end{restatable}

\begin{restatable}{assumption}{rotinputdiscrimination}
\label{assumption:rot_input}
    \textbf{(Discrimination.)} 
    A discriminative direction $\featuredeterminingdirection$ exists for $\inputfunc$ with threshold $\discriminationthreshold\!>\!0$.\looseness=-1
\end{restatable}

\begin{restatable}{assumption}{rotsufficientseparation}
\label{assumption:rot_separation}
    \textbf{(Sufficient separation.)} 
    Let $\steeringgap$, $\steeringmax$ and $\classspread$ be defined as in \cref{assumption:separation}.
    We require the discriminative gap to exceed the dynamic range of the steering projection: $\discriminationthreshold > \steeringmax + \steeringthreshold$.
\end{restatable}

\newcommand{\classspreadsd}{\varepsilon^{\mathtt{SD}}_{\mathtt{spread}}}

\begin{restatable}{assumption}{rotstability}
\label{assumption:rot_stability}
    \textbf{(Within-class stability.)} 
    Let $\classspreadsd$ denote the within-class variation along $\outputinfluencedirection$:
    \begin{align}
        \classspreadsd = \max_{i \in \{0,1\}} \max_{\hiddenstate_b, \hiddenstate_s \in \mathcal{H}_i}
            \left| \langle \hiddenstate_b - \hiddenstate_s, \outputinfluencedirection \rangle \right|.
    \end{align}
    We require $\classspread + \classspreadsd < 2 \steeringgap$.
\end{restatable}

\begin{restatable}{thm}{rotationthm}\label{theorem:rotation}
Let $\dnn$ be a DNN, and $\alg$ be an algorithm with $\algfuncnode{\nodeint} = \inputfunc$ and $\algfuncnode{\nodesoutput} = \outputfunc$.
Further, let $\Rotation$ be the class of orthogonal maps.
Under assumptions 4--7, $\alg$ is a $\Rotation$-abstraction of $\dnn$.
\end{restatable}

\newcommand{\projection}{\mathbf{P}}
\setcounter{mytheoremstep}{0}

\begin{proof}
    We will prove this theorem by construction, following the same steps as the proof of \cref{theorem:general}.
    More precisely, we will prove it in 4 steps:
    \begin{enumerate}
        \item First, we will construct an orthogonal alignment map $\causalmap \in \Rotation$---expanded as $\causalmap(\hiddenstate) = \projection\,\hiddenstate$ for an orthogonal $\projection \in \R^{d \times d}$---whose first row is
        \begin{align}
            \mathbf{v}_1 = \frac{\featuredeterminingdirection - \outputinfluencedirection}{\|\featuredeterminingdirection - \outputinfluencedirection\|} .
        \end{align}
        \item Second, we show that, given such an orthogonal matrix, our interventions have form:
        \begin{align}
            \hiddensrcbase
            = \hiddenbase + \langle \mathbf{v}_1, \hiddensrc - \hiddenbase \rangle \, \mathbf{v}_1 ,
        \end{align}
        Unlike the general case, orthogonality forces the read and write directions to coincide---both equal to $\mathbf{v}_1$.
        \item Third, we prove that, given Assumptions 4--7, within class interventions preserve the DNN's output. Formally, this shows that for these interventions $\bybasesrcN = \bybasesrcA = \outputfunc(\bxbase, \inputfunc(\bxsrc))$;
        \item Fourth, we prove that, given Assumptions 4--7, across class interventions change the DNN's output. Formally, this shows that for these interventions $\bybasesrcN = \bybasesrcA = \outputfunc(\bxbase, \inputfunc(\bxsrc))$.
    \end{enumerate}
    Together, steps 3 and 4 cover all pairs of inputs $\bxsrc, \bxbase$, showing that for every such pair $\bybasesrcN = \bybasesrcA$.
    The proof for each of these steps is given below, and together they prove our theorem.
    As in \cref{theorem:general}, our proof assumes that the data is centered around the origin along $\featuredeterminingdirection$ and $\outputinfluencedirection$. We do this without loss of generality, since the argument only depends on relative differences along $\featuredeterminingdirection$ and distance away from the steering threshold along $\outputinfluencedirection$.
\end{proof}
\begin{mytheoremstep}
    There exists an orthogonal matrix $\projection \in \R^{d \times d}$ whose first row is $\mathbf{v}_1 = (\featuredeterminingdirection - \outputinfluencedirection) / \|\featuredeterminingdirection - \outputinfluencedirection\|$.
\end{mytheoremstep}
\begin{proof}
    \Cref{assumption:rot_separation} implies $|\langle \featuredeterminingdirection, \outputinfluencedirection \rangle| < 1$: if $\featuredeterminingdirection = \pm\outputinfluencedirection$, then every $\hiddenstate \in \calH$ would have $|\langle\hiddenstate,\outputinfluencedirection\rangle| = |\langle\hiddenstate,\featuredeterminingdirection\rangle| > \discriminationthreshold$, giving $\steeringmax > \discriminationthreshold$, a contradiction. Hence $\featuredeterminingdirection$ and $\outputinfluencedirection$ are not collinear, $\featuredeterminingdirection - \outputinfluencedirection \neq \mathbf{0}$, and $\mathbf{v}_1$ is well defined.
    We can extend to an orthonormal basis $\{\mathbf{v}_1, \mathbf{v}_2, \mathbf{v}_3, \dots, \mathbf{v}_d\}$ of $\R^d$, and define $\projection$ as the change-of-basis matrix with rows $\mathbf{v}_1^\top, \mathbf{v}_2^\top, \dots, \mathbf{v}_d^\top$.
    As its rows form an orthonormal basis, $\projection$ is orthogonal, with its first row being $\mathbf{v}_1^\top$ by construction.
\end{proof}

\begin{mytheoremstep}
    Consider an orthogonal alignment map $\causalmap \in \Rotation$ where $\causalmap(\hiddenstate) = \projection\,\hiddenstate$ for an orthogonal matrix $\projection \in \R^{d\times d}$ with first row $\mathbf{v}_1^\top$.
    Given this alignment map, our interventions have form:
    \begin{align}
        \hiddensrcbase = \hiddenbase + \langle \mathbf{v}_1, \hiddensrc - \hiddenbase \rangle \, \mathbf{v}_1 .
    \end{align}
    Moreover, writing $d_b = \langle \hiddenbase, \featuredeterminingdirection \rangle$, $d_s = \langle \hiddensrc, \featuredeterminingdirection \rangle$, $\mathsf{s}_b = \langle \hiddenbase, \outputinfluencedirection \rangle$, and $\mathsf{s}_s = \langle \hiddensrc, \outputinfluencedirection \rangle$, the induced displacement along the steering direction is
    \begin{align}\label{eq:rotation_change}
        \langle \hiddensrcbase - \hiddenbase,\, \outputinfluencedirection \rangle
        = \tfrac{1}{2}(d_b - d_s) + \tfrac{1}{2}(\mathsf{s}_s - \mathsf{s}_b) .
    \end{align}
\end{mytheoremstep}
\begin{proof}
    As $\projection$ is orthogonal it is invertible, so the intervention factorization from Step~2 of the proof of \cref{theorem:general} applies: $\hiddensrcbase = \hiddenbase + \langle \mathbf{r}, \hiddensrc - \hiddenbase \rangle\, \mathbf{w}$, where $\mathbf{r}$ is the first row of $\projection$ and $\mathbf{w}$ is the first column of $\projection^{-1}$.
    For an orthogonal map, $\projection^{-1} = \projection^\top$, so the first column of $\projection^{-1}$ equals the first row of $\projection$: $\mathbf{w} = \projection^\top\basevector_1 = \mathbf{v}_1 = \mathbf{r}$.
    The read and write directions thus coincide, giving $\hiddensrcbase = \hiddenbase + \langle \mathbf{v}_1, \hiddensrc - \hiddenbase \rangle\, \mathbf{v}_1$.

    For the displacement along the steering direction, we take the inner product with $\outputinfluencedirection$:
    \begin{align}\label{eq:thm2_decomposition}
        \langle \hiddensrcbase - \hiddenbase,\, \outputinfluencedirection \rangle = \langle \mathbf{v}_1, \hiddensrc - \hiddenbase \rangle\,\langle \mathbf{v}_1, \outputinfluencedirection \rangle .
    \end{align}
    Using $\|\featuredeterminingdirection - \outputinfluencedirection\|^2 = \langle\featuredeterminingdirection,\featuredeterminingdirection\rangle - 2\,\langle\featuredeterminingdirection,\outputinfluencedirection\rangle + \langle\outputinfluencedirection,\outputinfluencedirection\rangle = 2(1 - \langle\featuredeterminingdirection,\outputinfluencedirection\rangle)$, we have:
    \begin{subequations}
    \begin{align}
        \langle \mathbf{v}_1, \outputinfluencedirection \rangle 
        &= \left\langle \frac{\featuredeterminingdirection - \outputinfluencedirection}{\|\featuredeterminingdirection - \outputinfluencedirection\|}, \outputinfluencedirection \right\rangle 
        & \mathcomment{definition of }\mathbf{v}_1 \\
        &= \frac{\langle\featuredeterminingdirection,\outputinfluencedirection\rangle - 1}{\|\featuredeterminingdirection - \outputinfluencedirection\|} \\
        % &= (\langle\featuredeterminingdirection,\outputinfluencedirection\rangle - 1)/\|\featuredeterminingdirection - \outputinfluencedirection\| \\
        &= -\frac{1}{2}\|\featuredeterminingdirection - \outputinfluencedirection\| 
        & \mathcomment{$\langle\featuredeterminingdirection,\outputinfluencedirection\rangle - 1 = -\frac{1}{2}\|\featuredeterminingdirection - \outputinfluencedirection\|^2$}
    \end{align}
    \end{subequations}
    % $\langle \mathbf{v}_1, \outputinfluencedirection \rangle = (\langle\featuredeterminingdirection,\outputinfluencedirection\rangle - 1)/\|\featuredeterminingdirection - \outputinfluencedirection\| = -\tfrac{1}{2}\|\featuredeterminingdirection - \outputinfluencedirection\|$ 
    Similarly, we have: 
    \begin{subequations}
    \begin{align}
        \langle \mathbf{v}_1, \hiddensrc - \hiddenbase \rangle 
        &= \left\langle \frac{\featuredeterminingdirection - \outputinfluencedirection}{\|\featuredeterminingdirection - \outputinfluencedirection\|}, \hiddensrc - \hiddenbase \right\rangle 
        & \mathcomment{definition of }\mathbf{v}_1 \\
        &= \frac{(d_s - \mathsf{s}_s) - (d_b - \mathsf{s}_b)}{\|\featuredeterminingdirection - \outputinfluencedirection\|} & \mathcomment{definitions}
    \end{align}
    \end{subequations}
    % $\langle \mathbf{v}_1, \hiddensrc - \hiddenbase \rangle = \big((d_s - \mathsf{s}_s) - (d_b - \mathsf{s}_b)\big)/\|\featuredeterminingdirection - \outputinfluencedirection\|$.
    Replacing both and multiplying them in \cref{eq:thm2_decomposition} leads to \cref{eq:rotation_change}.
\end{proof}

\begin{mytheoremstep}
    Let $\bxbase, \bxsrc \in \calX$ be two inputs sharing a class, i.e., $\inputfunc(\bxbase) = \inputfunc(\bxsrc)$.
    Further, let $\causalmap$ be any orthogonal alignment map with first row $\mathbf{v}_1^\top$ (which exists by Step~1).
    The DNN $\dnn$'s output is preserved under this intervention, meaning that $\bybasesrcN = \bybasesrcA = \outputfunc(\bxbase, \inputfunc(\bxsrc))$.
\end{mytheoremstep}
\begin{proof}
    Recall that $\steeringgap$ (\cref{assumption:rot_separation}) is the minimum gap between hidden states and the $\pm \steeringthreshold$ steering threshold.
    As $\bxbase, \bxsrc$ share a class, if we show the intervention does not make hidden states cross the $\steeringthreshold$ boundary, they will preserve their class and the condition will hold.
    In other words, we need to show that the displacement along the steering direction is smaller than $\steeringgap$:
    \begin{align}
        |\langle \hiddensrcbase - \hiddenbase,\, \outputinfluencedirection \rangle| < \steeringgap .
    \end{align}
    By \cref{eq:rotation_change} and the triangle inequality,
    \begin{subequations}
    \begin{align}
        |\langle \hiddensrcbase - \hiddenbase,\, \outputinfluencedirection \rangle|
        &= \left| \tfrac{1}{2}(d_b - d_s) + \tfrac{1}{2}(\mathsf{s}_s - \mathsf{s}_b) \right| \\
        &\leq \tfrac{1}{2}|d_s - d_b| + \tfrac{1}{2}|\mathsf{s}_s - \mathsf{s}_b| \\
        &\leq \tfrac{1}{2}\classspread + \tfrac{1}{2}\classspreadsd
        & \mathcomment{definitions of $\classspread$ and $\classspreadsd$} \\
        &< \steeringgap
        & \mathcomment{$\classspread + \classspreadsd < 2\steeringgap$}
    \end{align}
    \end{subequations}
    which completes this step.
\end{proof}

\begin{mytheoremstep}
    Let $\bxbase, \bxsrc \in \calX$ be two inputs with opposite classes, i.e., $\inputfunc(\bxbase) \neq \inputfunc(\bxsrc)$.
    Further, let $\causalmap$ be any orthogonal alignment map with first row $\mathbf{v}_1^\top$ (which exists by Step~1).
    The DNN $\dnn$'s output flips to the source's class under this intervention, meaning that $\bybasesrcN = \bybasesrcA = \outputfunc(\bxbase, \inputfunc(\bxsrc))$.
\end{mytheoremstep}
\begin{proof}
    Recall that $\steeringmax$ (\cref{assumption:rot_separation}) is the maximum activation of the steering direction by hidden states.
    As $\bxbase, \bxsrc$ have opposite classes, if we show the intervention moves the hidden state across the $\steeringthreshold$ boundary in the correct direction, they will force the output class to change correctly.
    Unlike the general case, an orthogonal map cannot amplify the displacement, so instead we bound the intervened activation $\langle \hiddensrcbase, \outputinfluencedirection \rangle$ directly, using the absolute separation $\discriminationthreshold > \steeringmax + \steeringthreshold$ of \cref{assumption:rot_separation}.

    Assume $\inputfunc(\bxbase)=0$ and $\inputfunc(\bxsrc)=1$, so $\hiddenbase \in \mathcal{H}_0$ and $\hiddensrc \in \mathcal{H}_1$.
    By \cref{assumption:rot_input}, $d_b < -\discriminationthreshold$ and $d_s > \discriminationthreshold$, so $d_b - d_s < -2\discriminationthreshold$.
    Adding $\mathsf{s}_b = \langle \hiddenbase, \outputinfluencedirection\rangle$ to \cref{eq:rotation_change} and using $\mathsf{s}_b, \mathsf{s}_s \leq \steeringmax$:
    \begin{subequations}
    \begin{align}
        \langle \hiddensrcbase,\, \outputinfluencedirection \rangle
        &= \tfrac{1}{2}(d_b - d_s) + \tfrac{1}{2}(\mathsf{s}_b + \mathsf{s}_s) \\
        &< -\discriminationthreshold + \steeringmax
        & \mathcomment{$d_b - d_s < -2\discriminationthreshold$ and $\mathsf{s}_b, \mathsf{s}_s \leq \steeringmax$} \\
        &< -\steeringthreshold
        & \mathcomment{$\discriminationthreshold > \steeringmax + \steeringthreshold$}
    \end{align}
    \end{subequations}
    Thus $\hiddensrcbase$ is moved past the $-\steeringthreshold$ boundary, flipping the output regardless of where $\hiddenbase$ originally resided along $\outputinfluencedirection$.
    For $\inputfunc(\bxbase)=1$ and $\inputfunc(\bxsrc)=0$, the same argument with signs reversed gives $\langle \hiddensrcbase, \outputinfluencedirection \rangle > \steeringthreshold$.
    This completes the proof.
\end{proof}

\section{Training Details}\label{app:training_details}
We use the \texttt{pyvene} \citep{wu-etal-2024-pyvene} package to implement and train alignment maps for both \pa and \co.

We parameterize our alignment maps of class $\RotationLinear$ as compositions of a $\Rotation$ map with a $\Linear$ map. In practice, the linear part of our map is parameterized as $U D U^\top$, where $U \in \mathbb{R}^{i \times i}$ is orthogonal and $D$ is diagonal with strictly positive entries. This convenient choice of alignment map, which is strictly contained in the broader $\RotationLinear$, turns out to be sufficient for discovering parasite features.

Our alignment maps are randomly initialized and optimized through backpropagation using a standard cross-entropy loss between the `correct' label under intervention and the actual output of the model under intervention. The model's weights are frozen during this process.

In the case of \pa, we train DAS on 400 unique, balanced, minimal pairs, and evaluate on a held-out test set of 50 minimal pairs \citep[as per][]{arora-etal-2024-causalgym}. We do so with 5 epochs with a batch size of 4 with a learning rate of 5e--3. Different from \citet{arora-etal-2024-causalgym}, our training set includes pairs which do not differ in input feature, with the intended intervention behavior as no change in \textsc{\textbf{Odds}}. 

For \co, we train DAS on 34,400 unique minimal pairs for 20 epochs with a batch size of 64 with a learning rate of 1e--4, and evaluate the accuracy of our interventions on a held-out test set of 1000 items \citep[as per][]{mib-2025}. 

All training is done on A40 GPUs. For \pa, training time for 5 epochs on 1 layer of \texttt{pythia-1b} \citep{biderman2023pythia} is 2 minutes on an A40. For \co, training time for 20 epochs on 1 layer of \texttt{Llama3.1-8b} \citep{Dubey2024TheL3} is 95 minutes.

\section{Parasite Features in Other Models}
\label{app:other_models}
To determine whether $\RotationLinear$'s ability to find parasite features is model-specific, we also compare its performance on \pa against $\Rotation$ in two other models. Specifically, we perform the same experiments as in \Cref{sec:set-up}, but with two models of different families and different sizes: \texttt{GPT2-Medium} \citep{radford2019language} and \texttt{Qwen2-1.5b} \citep{yang2024qwen2technicalreport}. 

Our results can be seen in \Cref{tab:other_models}. Notably, we see that, in line with our previous results, maps of the class \rotationlinearname can find parasite features for \pa more so than those of rotation in \texttt{Qwen2-1.5b}. That is, both the maximal \textbf{\textsc{Odds}} across layers is higher for \rotationlinearname than for rotation, and \rotationlinearname shows higher causal efficacy across all layers. Interestingly, we find that each intervention is roughly equal in \texttt{GPT2-Medium}. This could be due to many causes, such as rotations finding parasites in this model or parasites being a phenomenon that emerges with scale. However, we do not attempt to adjudicate on such questions here, instead leaving it up to future work.\looseness=-1

\begin{table}[h]
\centering
\footnotesize
\setlength{\tabcolsep}{8pt}            % more room between columns
\renewcommand{\arraystretch}{1.2}      % taller rows
\setlength{\aboverulesep}{0pt}
\setlength{\belowrulesep}{0pt}
\rowcolors{4}{white}{gray!7}
\begin{tabular}{c cc cc}
\toprule
\rowcolor{gray!25}
\rule{0pt}{3ex}% extra height in the first header row
 & \multicolumn{2}{c}{\textbf{\textit{GPT2-medium}}} & \multicolumn{2}{c}{\textbf{\textit{Qwen2-1.5b}}} \\
\cmidrule(lr){2-3} \cmidrule(lr){4-5}
\rowcolor{gray!12}
\textbf{Layer} & $\Rotation$ & $\RotationLinear$ & $\Rotation$ & $\RotationLinear$ \\
\midrule
0  & 2.43 & 2.38 & 0.11 & 0.24 \\
1  & \textbf{2.53} & \textbf{2.46} & 1.59 & 2.03 \\
2  & 2.46 & 2.33 & 1.59 & 1.99 \\
3  & 2.35 & 2.25 & 1.85 & 2.04 \\
4  & 1.89 & 1.80 & \textbf{2.03} & 2.05 \\
5  & 1.64 & 1.47 & 1.81 & 2.22 \\
6  & 1.67 & 1.30 & 0.98 & 1.89 \\
7  & 1.72 & 1.37 & 1.53 & 1.98 \\
8  & 1.65 & 1.28 & 1.29 & 1.82 \\
9  & 1.71 & 1.39 & 1.64 & 2.07 \\
10 & 1.69 & 1.31 & 1.45 & 2.21 \\
11 & 1.56 & 1.43 & 1.51 & \textbf{2.83} \\
12 & 2.03 & 2.06 & 1.53 & 2.55 \\
13 & 2.13 & 0.92 & 1.85 & 2.43 \\
14 & 2.31 & 1.08 & 1.80 & 2.55 \\
15 & 1.96 & 0.94 & 1.40 & 2.23 \\
16 & 1.67 & 0.77 & 1.67 & 2.06 \\
17 & 1.44 & 0.70 & 1.69 & 2.19 \\
18 & 0.76 & 0.29 & 1.92 & 2.39 \\
19 & 0.52 & 0.50 & 1.93 & 2.26 \\
20 & 0.18 & 0.21 & 1.66 & 2.06 \\
21 & 0.18 & 0.16 & 1.74 & 2.09 \\
22 & 0.10 & 0.11 & 1.52 & 1.92 \\
23 & 0.00 & 0.00 & 1.30 & 1.70 \\
24 & --   & --   & 0.15 & 0.37 \\
25 & --   & --   & 0.16 & 0.13 \\
26 & --   & --   & 0.04 & 0.09 \\
27 & --   & --   & 0.00 & 0.00 \\
\bottomrule
\end{tabular}
\vspace{6pt}                           % space before caption
\caption{\pa with 1D interventions using rotation and \rotationlinearname maps. Bold marks the best layer for each map. GPT2-medium has 24 layers.}
\label{tab:other_models}
\end{table}

\newpage

\end{document}